\documentclass[lettersize,journal]{IEEEtran}
\usepackage{amsmath,amssymb,amsfonts}
\usepackage{algorithmic}
\usepackage{graphicx}
\usepackage{textcomp}
\usepackage{xcolor}
\usepackage{booktabs}
\usepackage{subfigure}
\usepackage{multirow}
\usepackage{hyperref}
\usepackage{algorithmic}
\usepackage{listings}
\usepackage{algorithm}

\begin{document}

\title{Reliable Node Similarity Matrix Guided Contrastive Graph Clustering}



\author{Yunhui~Liu, 
        Xinyi~Gao, 
        Tieke~He*,~\IEEEmembership{Member,~IEEE}, 
        Tao~Zheng, 
        Jianhua~Zhao, and\\
        Hongzhi~Yin,~\IEEEmembership{Senior Member,~IEEE}
        \IEEEcompsocitemizethanks{
        \IEEEcompsocthanksitem Yunhui~Liu, Tieke~He, Tao~Zheng and Jianhua~Zhao are with the State Key Laboratory for Novel Software Technology, Nanjing University, Nanjing 210023, China (email: \{lyhcloudy1225, hetieke\}@gmail.com; \{zt, zhaojh\}@nju.edu.cn). 
        \IEEEcompsocthanksitem Xinyi~Gao and Hongzhi~Yin are with the School of Information Technology and Electrical Engineering, The University of Queensland, Brisbane, Australia (email: \{xinyi.gao, h.yin1\}@uq.edu.au).}
        \thanks{*Corresponding author.}}




\maketitle

\begin{abstract}
Graph clustering, which involves the partitioning of nodes within a graph into disjoint clusters, holds significant importance for numerous subsequent applications. Recently, contrastive learning, known for utilizing supervisory information, has demonstrated encouraging results in deep graph clustering. This methodology facilitates the learning of favorable node representations for clustering by attracting positively correlated node pairs and distancing negatively correlated pairs within the representation space. Nevertheless, a significant limitation of existing methods is their inadequacy in thoroughly exploring node-wise similarity. For instance, some hypothesize that the node similarity matrix within the representation space is identical, ignoring the inherent semantic relationships among nodes. Given the fundamental role of instance similarity in clustering, our research investigates contrastive graph clustering from the perspective of the node similarity matrix. We argue that an ideal node similarity matrix within the representation space should accurately reflect the inherent semantic relationships among nodes, ensuring the preservation of semantic similarities in the learned representations. In response to this, we introduce a new framework, Reliable Node Similarity Matrix Guided Contrastive Graph Clustering (NS4GC), which estimates an approximately ideal node similarity matrix within the representation space to guide representation learning. Our method introduces node-neighbor alignment and semantic-aware sparsification, ensuring the node similarity matrix is both accurate and efficiently sparse. Comprehensive experiments conducted on $8$ real-world datasets affirm the efficacy of learning the node similarity matrix and the superior performance of NS4GC. 
The implementation code can be found at: \url{https://github.com/Cloudy1225/NS4GC}.
\end{abstract}


\section{Introduction}
Clustering is significantly important in numerous practical applications and plays a crucial role in the field of machine learning. One specific task, graph clustering, involves partitioning nodes within a graph into disjoint groups. Graph clustering has demonstrated its effectiveness in various domains, including social network analysis, recommender systems, bioinformatics, and medical science \cite{DGCSurvey, GSSLSurvey}. Recently, the emergence of deep graph clustering, which adeptly captures both structural relationships and node attribute information, signifies a notable trend in the evolution of graph clustering methods \cite{VGAE, DGI, MAGC, CCGC, RCAGL}.

A prominent category of deep graph clustering methods is comprised of Graph Contrastive Learning (GCL)-based approaches \cite{MVGRL, SCGC, CCGC, MLG-CPC}, which aim to learn a representation invariant to augmentations of the same graph instance. Generally, GCL-based methods can be divided into two groups, each following a two-step strategy \cite{DGCSurvey}. Initially, graph contrastive learning is employed in the pre-training phase to yield informative node representations. Subsequently, the first group \cite{SCGC, CCGC} utilizes conventional clustering techniques, such as k-means, on the obtained representations. The second group \cite{MLG-CPC} refines the learned representations using cluster-oriented pseudo-supervision, aiming to concurrently perform clustering and embedding learning. Given the foundational role of representation learning in the pre-training phase for all GCL-based methods, our work focuses on learning representations more conducive to clustering.

\begin{figure*}[!ht]
    \centering
    \subfigure[Input Graph]{\includegraphics[width=0.16\linewidth]{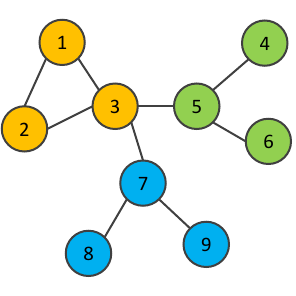}}   
    \subfigure[Input Adjacency]{\includegraphics[width=0.16\linewidth]{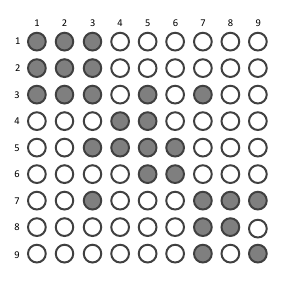}}
    \subfigure[Ideal NSM]{\includegraphics[width=0.16\linewidth]{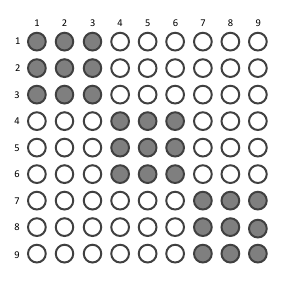}}
    \subfigure[GRACE's NSM]{\includegraphics[width=0.16\linewidth]{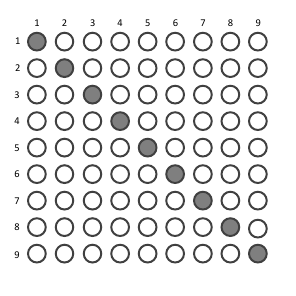}}
    \subfigure[Our NSM]{\includegraphics[width=0.16\linewidth]{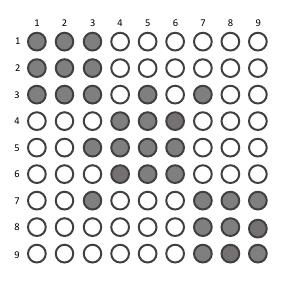}}
    \subfigure[Our Adjacency]{\includegraphics[width=0.16\linewidth]{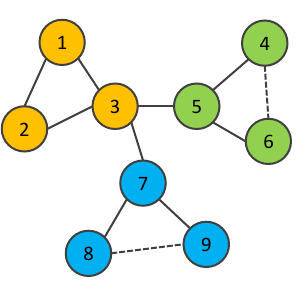}}
    
    \caption{Visualization of the node similarity matrix (NSM), with $\bullet$/$\circ$ denoting the pairwise semantic similarity/difference (1/0). (a) depicts the input attributed graph, where the digit is the node index and the color is the class index. (b) represents the input adjacency matrix, serving as a noise and incomplete surrogate for the ideal node similarity matrix. (c) is the ideal node similarity matrix. (d) illustrates the node similarity matrix for node-node contrastive learning methods, such as GRACE \cite{GRACE}, where it is implicitly assumed to be an identity matrix. (e) displays our learned node similarity matrix, characterized as a refined adjacency matrix, enriched with additional intra-cluster edges, as depicted in (f). For intra-cluster but disconnected node pairs, like $(4,6)$, they are expected to exhibit relatively high cosine similarity due to their inherently similar input features or the influence of the message passing mechanism in graph neural networks. In this case, node-node contrastive methods such as GRACE impose heavier penalties on such pairs to distance them. In contrast, our semantic-aware sparsification mitigates heavy penalization for them, thereby preserving their high similarity to a considerable extent (see Gradient Analysis in Section \ref{Sec: Gradient Analysis}).}
    \label{Fig: NSM}
  \end{figure*}

Clustering algorithms, such as k-means, typically rely on distance or similarity measures between pairs of instances to group semantically similar instances within the same cluster \cite{Clustering}. Therefore, the ideal representation space for graph clustering is one where semantically similar nodes (i.e., nodes within the same ground-truth cluster) exhibit higher similarity than semantically different nodes (i.e., nodes from different ground-truth clusters). To provide visual clarity, we depict the ideal node similarity matrix within the representation space in Figure \ref{Fig: NSM}(c). For simplicity, the node similarity matrix takes the form of a binary matrix, indicating binary semantic similarity between any pair of nodes in the graph and demonstrating a sparse structure. Specifically, similarities corresponding to nodes within the same ground-truth cluster are denoted as $1$, while similarities corresponding to nodes from different ground-truth clusters are denoted as $0$. Ideally, if we could obtain such a node similarity matrix to guide the representation learning process, resulting node representations would effectively preserve semantic similarities between nodes and prove highly conducive to clustering. However, obtaining the ideal node similarity matrix is generally unfeasible due to the absence of labels or supervision, rendering it impractical to employ the node similarity matrix directly for representation learning.

We further undertake a study of contrastive graph clustering from the perspective of learning the node similarity matrix. Node-node contrastive methods, such as GRACE \cite{GRACE}, GCA \cite{GCA} implicitly assume the node similarity matrix to be an identity matrix. These methods aim to align positive pairs, which involve augmented node pairs derived from the same node, while simultaneously pushing apart negative pairs, encompassing any two distinct nodes from the training graph. Essentially, only the diagonal entries of the node similarity matrix within the representation space are set to $1$, as depicted in Figure \ref{Fig: NSM}(d). In contrast, some negative-free methods, like CCASSG \cite{CCA-SSG} and BGRL \cite{BGRL}, focus solely on aligning positive pairs, which can potentially result in a trivial solution where all representations collapse into a single point, i.e., the node similarity matrix within the representation space becomes an all-one matrix. Therefore, the node-node contrastive methods and the negative-free methods represent two opposing extremes: node-node contrastive methods assume the node similarity matrix to be an identity matrix (or extremely sparse), while negative-free methods do not utilize any sparse structure within the node similarity matrix.

To mitigate this issue and achieve a balanced approach, we strive to estimate an approximately ideal node similarity matrix within the representation space to guide the representation learning phase. This matrix is anticipated to be both reliable and suitably sparse. Our investigation thus explores the principle of homophily in real-world graphs, which suggests that connected nodes tend to share similar underlying semantics \cite{Homophily}, thereby mirroring prevalent cluster information within actual graph structures. This phenomenon is substantiated by the high node homophily figures in Table \ref{Tab: Dataset statistics} and the enhanced clustering outcomes presented in Table \ref{Tab: Clustering Performance}. Specifically, we begin by aligning nodes with their neighbors in the representation space, enhancing the similarity of connected nodes. This approach sets entries for connected node pairs in the node similarity matrix to $1$. Unlike node-node contrastive methods that predict sparse ($0$) entries for unconnected node pairs, we employ a semantic-aware sparsification technique. This technique fine-tunes the level of sparsity in the node similarity matrix, retaining entries close to $1$ for node pairs with relatively high semantic similarity and setting others to $0$. Hence, our refined node similarity matrix essentially acts as an enhanced adjacency matrix that includes more intra-cluster connections, as illustrated in Figures \ref{Fig: NSM}(e) and (f). Consequently, we introduce a novel contrastive graph clustering method, NS4GC. NS4GC utilizes the refined node similarity matrix to discern and leverage underlying semantic similarities among nodes for clustering purposes. Extensive experiments on eight real-world datasets validate the effectiveness of our approach in learning a reliable node similarity matrix and demonstrate the superior performance of the proposed NS4GC method.

The contributions of this work can be summarized as follows:
\begin{itemize}
    \item \textbf{New Problem and Insights.} We investigate contrastive graph clustering from the perspective of the node similarity matrix, and identify crucial issues in both node-node contrastive methods, which assume identical node similarity matrices, and negative-free methods associated with excessively dense matrices.

    \item \textbf{New Methodology.} We introduce a new clustering-friendly contrastive learning method that utilizes an approximated ideal node similarity matrix to guide the representation learning process. This method integrates node-neighbor alignment and semantic-aware sparsification, ensuring the node similarity matrix is both dependable and suitably sparse.

    \item \textbf{SOTA Performance.} Through comprehensive experimentation on eight benchmark datasets spanning various fields, our method demonstrates its ability to learn a more accurate node similarity matrix within the representation space and outperforms various state-of-the-art (SOTA) methods.
\end{itemize}

\section{Related Work}

\subsection{Graph Neural Networks}
GNNs comprise a group of neural network models adept at capturing both graph structure and node attribute information \cite{GCN, GAT, GraphSAGE, APPNP}. They serve the purpose of graph representation learning and facilitate various tasks, such as recommender systems \cite{BiHGH, XSimGCL}, anomaly detection \cite{BOURNE}, molecular graph generation \cite{Graphusion}, and graph distillation \cite{MCond, GCSurvey}. Notably, GCN \cite{GCN} stands out as one of the most influential models, which extends the conventional convolution operation, originally designed for sequential or grid data, to graph-structured data. Furthermore, GAT \cite{GAT} introduces an attention mechanism, enabling a more flexible aggregation of messages from neighbors by learning the importance of each such neighbor. On the other front, GraphSAGE \cite{GraphSAGE} adopts a sampling-based approach to gather neighbor information, making it particularly suitable for large-scale graphs. APPNP \cite{APPNP} breaks new ground by decoupling prediction and propagation, effectively addressing the inherent limited-range issue present in many neighbor aggregation models. For most deep graph clustering methods, GNN plays a pivotal role in embedding nodes into a lower-dimensional space. Similar to most previous works \cite{GRACE, CCA-SSG, BGRL}, we adopt GCN as the foundational graph encoder.

\subsection{Attributed Graph Clustering}
Recently, substantial methods have emerged in attributed graph clustering, largely owing to their strong capacity to represent both attribute and structural information within graphs \cite{DGCSurvey}. The majority of attributed graph clustering methods adopt a two-stage approach, conducting traditional or neural clustering subsequent to obtaining low-dimensional node representations. Notably, among these methods, Graph Contrastive Learning (GCL) and Graph Auto-Encoder (GAE) stand out as highly effective strategies for representation learning.

\subsubsection{GCL-based Clustering Methods} \label{Sec: Discussion}
Contrastive graph clustering methods employ a crucial concept of enhancing the discriminability of features by attracting positive graph instances and repelling negative ones \cite{DGCSurvey}. For instance, DGI \cite{DGI} and MVGRL \cite{MVGRL} employ the mutual information maximization principle to contrast node-level representations with (sub-)graph-level representations. On the other hand, node-node contrastive methods GRACE \cite{GRACE} and GCA \cite{GCA} adopt InfoNCE \cite{InfoNCE} as their objective, aiming to align augmented node pairs originating from the same node while pushing apart any two distinct nodes from the training graph. gCooL \cite{gCooL} further integrates community detection and InfoNCE to capture cluster-wise similarity. Negative-free methods CCASSG \cite{CCA-SSG} and BGRL \cite{BGRL} focus primarily on aligning positive pairs, preventing representation collapse through the use of asymmetric network architecture and regularization of the empirical covariance matrix of representations, respectively. Additionally, SCGC \cite{SCGC} simplifies graph augmentation through parameter-unshared Siamese encoders and embedding disturbance, while CCGC \cite{CCGC} taps into intrinsic supervision information from high-confidence clustering results to enhance the quality of positive and negative instances.

\textbf{Discussion.} 
Here we analyze the methods above from the perspective of the node similarity matrix, since computing node distance/similarity is essential for clustering. An ideal node similarity matrix should reflect the inherent semantic similarity/difference between any pair of nodes within the graph. Guided by this node similarity matrix, the learned representations of semantically similar nodes are positioned closely within the representation space, thereby facilitating the clustering of such nodes into cohesive groups. However, in practice, GRACE and GCA implicitly assume the node similarity matrix to be an identity matrix, while BGRL and CCASSG lack explicit constraints on off-diagonal entries. DGI and MVGRL, which contrast node-level representations with (sub-)graph-level representations, yield a highly uncertain node similarity matrix. gCooL incorporates a community/clustering detection module and attempts to align nodes with their respective community/cluster, but the accuracy of the detected communities/clusters is disappointing, introducing additional noise \cite{gCooL}. On the other hand, while CCGC can better capture node-wise similarity by carefully selecting true positive instances and hard negative instances, they require time-consuming k-means clustering at each iteration. In contrast, our method, utilizing a simple node-neighbor alignment and semantic-aware sparsification, can effectively learn an approximately ideal node similarity matrix, resulting in clustering-friendly representations.

\subsubsection{GAE-based Clustering Methods}
GAE/VGAE \cite{VGAE} is the pioneering graph auto-encoder model, which embeds node attributes along with structural information through a graph encoder, followed by the reconstruction of the graph adjacency using an inner product decoder. Following this, ARGA/ARVGA \cite{ARVGA} focuses on aligning latent representations with a prior distribution through adversarial learning. To address challenges associated with isolated nodes, GNAE and VGNAE \cite{GNAE} employ $\ell_2$ normalization, which effectively prevents representations from converging towards zero. These methods adopt a direct approach by applying k-means to pre-trained representations to obtain cluster assignments. In contrast, the subsequent methods refine the pre-trained representations through cluster-oriented pseudo supervision. DAEGC \cite{DAEGC}, for instance, aims to enhance clustering-oriented features by minimizing joint clustering and reconstruction losses after pre-training. Both SDCN \cite{SDCN} and DFCN \cite{DFCN} introduce a unified framework that jointly trains an auto-encoder and a graph auto-encoder. They mitigate the issue of oversmoothing by incorporating an information transport operation and a structure-attribute fusion module, respectively. More recently, EGAE \cite{EGAE} introduces an integration of relaxed k-means for improved clustering performance, while R-GAE \cite{RGAE} enhances GAE-based approaches by tackling feature randomness and feature drift.

\subsubsection{Community Clustering Methods}
Some studies have integrated traditional community detection techniques with graph neural networks to advance deep graph clustering. These efforts involve directly optimizing graph clustering by employing community-oriented loss functions, derivatives of well-established cohesiveness metrics. For instance, NOCD \cite{NOCD} introduces an overlapping community detection loss function anchored on maximizing the likelihood of Bernoulli-Poisson models \cite{BigCLAM}. Similarly, \cite{MinCutPool} devises a continuous relaxation of the normalized minCUT problem \cite{MinCut}, training a GNN to infer cluster assignments that minimize this criterion. Drawing inspiration from the modularity \cite{Modularity} clustering quality measure, \cite{DMoN} unveiled Deep Modularity Networks (DMoN) to extract high-quality clusters. Nonetheless, it is essential to note that these approaches inherit the limitations of the underlying metrics they seek to optimize, resulting in cluster assignments that are subject to the loss objective’s definition of cluster (community).

\subsubsection{Non-neural Clustering Methods}
While traditional k-means and spectral clustering \cite{SC} fail to simultaneously utilize both graph topology and node attributes, several recent non-neural approaches have addressed this limitation by drawing inspiration from graph signal processing \cite{GSP}. For instance, AGC \cite{AGC} conducts graph convolution using a low-pass graph filter, aiming to retain low-frequency basis signals and eliminate high-frequency ones within node attributes. SGC \cite{SGC} aggregates the information from long-range neighbors by applying the $K$-th power of the normalized adjacency matrix to better capture global cluster structures. SSGC \cite{SSGC} employs the Markov diffusion kernel to incorporate larger neighborhoods compared to SGC and cope better with oversmoothing. Notably, these ``non-neural" methods only propagate features along edges without employing neural networks for feature transformation, resulting in a representation space with the same dimensionality as the input space. To acquire cluster assignments, AGC utilizes spectral clustering on the learned node similarity matrix, while SGC and SSGC apply k-means on the representations after dimensionality reduction through truncated singular value decomposition.

\section{Preliminary}

\begin{table}
    \begin{center}
    {\caption{ Notation summary.}\label{Tab:  Notation summary}}
    \begin{tabular}{ll}
    \toprule
    Notation                      & Description      \\ \midrule 
    $\mathcal{G}$                 & attributed graph \\
    $\mathcal{V}, \mathcal{E}$    & node set and edge set \\
    $\boldsymbol{A}, \boldsymbol{X}$              & adjacency matrix and attribute matrix \\
    $\boldsymbol{I}, \boldsymbol{D}$              & identity matrix and diagonal degree matrix \\
    $\boldsymbol{\widetilde{A}}, \boldsymbol{\widetilde{X}}$  & augmented adjacency matrix and attribute matrix \\
    $\boldsymbol{Z}, \boldsymbol{C}$                          & node representations and cluster assignments matrix \\
    $\boldsymbol{W}, \boldsymbol{S}$ & node similarity matrix, cross-view cosine similarity matrix\\ \midrule
    $f_\theta$          & graph encoder with parameters $\theta$ \\
    $p_d, p_m$          & intensity of edge dropping and feature masking \\
    $s, \tau$           & split cosine similarity and temperature parameter \\
    $\lambda, \gamma$   & weights to balance the contributions of three loss items \\
    
    \bottomrule
    \end{tabular}
    \end{center}
\end{table}


\subsection{Problem Statement}
Let $\mathcal{G} = (\mathcal{V}, \mathcal{E})$ represent an attributed graph, where $\mathcal{V} = \{ v_1, v_2, \cdots, v_n\}$ denotes the node set, and $\mathcal{E} \subseteq \mathcal{V} \times \mathcal{V}$ comprises the edge set. This graph is associated with a feature matrix $\boldsymbol{X} \in \mathbb{R}^{n \times p}$, where each row $\boldsymbol{x}_i \in \mathbb{R}^p$ corresponds to the features of node $v_i$. Additionally, an adjacency matrix $\boldsymbol{A} \in \{ 0,1 \}^{n \times n}$ is defined, with $\boldsymbol{A}_{ij} = 1$ if and only if an edge exists between nodes $v_i$ and $v_j$. 

The objective of deep graph clustering is to encode nodes with a neural network in an unsupervised manner and then partition them into several disjoint groups. Generally, a neural network $f_\theta$ (often GCN \cite{GCN}) is initially trained without human annotations and embeds the nodes into the latent space by leveraging the node attributes and the graph structure: $\boldsymbol{Z} = f_\theta \left( \boldsymbol{A}, \boldsymbol{X} \right)$. Here, $\boldsymbol{Z} \in \mathbb{R}^{n \times d}$ represents the learned node representations. Subsequently, a clustering algorithm $\mathcal{C}$, such as k-means, spectral clustering \cite{SC}, or a clustering neural network layer \cite{SDCN}, is employed to partition nodes into $k$ disjoint groups: $\boldsymbol{C} = \mathcal{C}(\boldsymbol{Z})$, where $\boldsymbol{C} \in \mathbb{R}^{n \times k}$ denotes the cluster membership matrix for all $n$ nodes.

\subsection{Graph Convolutional Network}
Graph Convolutional Network (GCN) \cite{GCN} is one of the most popular graph neural networks. Mathematically, the formulation of the graph convolutional layer can be expressed as:
\begin{equation}
    \boldsymbol{Z}^{(l+1)} = \sigma(\boldsymbol{\hat{D}}^{-1/2} \boldsymbol{\hat{A}} \boldsymbol{\hat{D}}^{-1/2} \boldsymbol{Z}^{(l)} \boldsymbol{\Theta}^{(l)}),
\end{equation}
where $\sigma$ is the activation function, $\boldsymbol{\hat{A}} = \boldsymbol{A}+\boldsymbol{I}$ signifies the adjacency matrix with inserted self-loops, and $\boldsymbol{\hat{D}}_{ii} = \sum_{j=0} \boldsymbol{\hat{A}}_{ij}$ corresponds to the diagonal degree matrix. Moreover, $\boldsymbol{Z}^{(l)}$ denotes the node representations of the $l$-th hidden layer, while $\boldsymbol{\Theta}^{(l)}$ represents a trainable weight matrix.

\subsection{Graph Homophily}
Graph homophily suggests that connected nodes often share the same cluster/class, which serves as valuable prior knowledge in real-world graphs such as citation networks, co-purchase networks, or friendship networks \cite{Homophily}. A well-used metric for quantifying the homophily of a graph is node homophily, which measures the average proportion of neighbors with the same cluster/class as each node:
\begin{equation}
    \text{Homo} = \frac{1}{|\mathcal{V}|}\sum_{v \in \mathcal{V}} \frac{|{ u \in \mathcal{N}(v):c_v=c_u }|}{|\mathcal{N}(v)|},
\end{equation}
where $\mathcal{N}(v)$ denotes the neighbor set of node $v$, and $c_v$ represents the cluster/class to which node $v$ belongs. Table \ref{Tab: Dataset statistics} presents the node homophily values for eight benchmark datasets. It is noteworthy that the node homophily surpasses the accuracy of clustering results for all datasets (refer to Table \ref{Tab: Clustering Performance}).

\begin{figure*}
\centerline{\includegraphics[width=1.\linewidth]{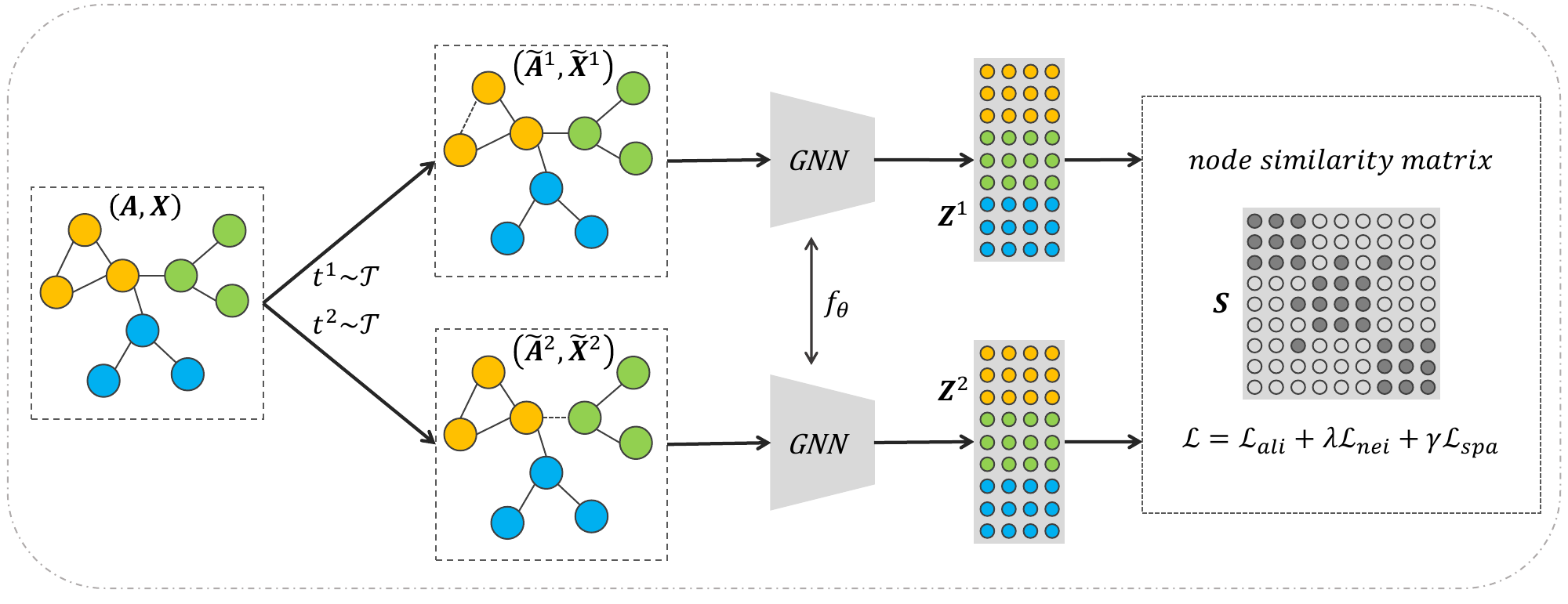}}
\caption{Overview of our proposed contrastive graph clustering framework NS4GC. For a given attributed graph, we first generate two distinct views via random augmentations: edge dropping and feature masking. These two views are subsequently fed into a shared GNN encoder to extract node representations. Then we instantiate the latent node similarity matrix using the cross-view cosine similarity. To optimize the model, we employ a combination of self-alignment loss, node-neighbor alignment loss, and a sparsity loss applied to the estimated node similarity matrix.} \label{Fig: Overview}
\end{figure*}

\section{Methodology}
We first describe the framework of the proposed method. Subsequently, we present how to learn an approximately ideal node similarity matrix within the representation space for clustering.

\subsection{Model Framework}
Our model simply comprises three components: 1) a random graph augmentation generator $\mathcal{T}$. 2) a GNN-based graph encoder $f_\theta$. 3) a novel objective function aiming to learn an approximately ideal node similarity matrix within the representation space. Figure \ref{Fig: Overview} is an illustration of the proposed model.

\subsubsection{Graph Augmentation}
The augmentation of graph data is a critical component of graph contrastive learning, as it generates diverse graph views, resulting in more generalized representations that are robust against variance. Following previous works \cite{GRACE, CCA-SSG, BGRL, BLNN}, we jointly adopt two widely utilized strategies, feature masking and edge dropping, to enhance graph attributes and topology information, respectively. 

\textbf{Feature Masking.}\quad
We randomly select a portion of the node features' dimensions and mask their elements with zeros. Formally, we first sample a random vector $\boldsymbol{\widetilde{m}} \in \{ 0, 1 \}^F$, where each dimension is drawn from a Bernoulli distribution with probability $1 - p_m$, i.e., $\widetilde{m}_i \sim \mathcal{B}(1 - p_m), \forall i$. Then, the masked node features $\widetilde {\boldsymbol{X}}$ are computed by $\parallel_{i=1}^N \boldsymbol{x}_i \odot \boldsymbol{\widetilde{m}}$, where $\odot$ denotes the Hadamard product and $\parallel$ represents the stack operation (i.e., concatenating a sequence of vectors along a new dimension).

\textbf{Edge Dropping.}\quad
In addition to feature masking, we stochastically drop a certain fraction of edges from the original graph. Formally, since we only remove existing edges, we first sample a random masking matrix $\boldsymbol{\widetilde{M}} \in \{ 0, 1 \}^{N \times N}$, with entries drawn from a Bernoulli distribution $\boldsymbol{\widetilde{M}}_{i,j} \sim \mathcal{B}(1 - p_d)$ if $\boldsymbol{A}_{i,j} = 1$ for the original graph, and $\boldsymbol{\widetilde{M}}_{i,j} = 0$ otherwise. Here, $p_d$ represents the probability of each edge being dropped. The corrupted adjacency matrix can then be computed as $\boldsymbol{\widetilde{A}} = \boldsymbol{A} \odot \boldsymbol{\widetilde{M}}$.


\subsubsection{Encoder Training}
During each training epoch, we first select two random augmentation functions, $t^1 \sim \mathcal{T}$ and $t^2 \sim \mathcal{T}$, where $\mathcal{T}$ is composed of all the possible graph transformation operations. Subsequently, two different views, $(\boldsymbol{\widetilde{A}}^1, \boldsymbol{\widetilde{X}}^1) = t^1(\boldsymbol{A},\boldsymbol{X})$ and $(\boldsymbol{\widetilde{A}}^2, \boldsymbol{\widetilde{X}}^2)=t^2(\boldsymbol{A},\boldsymbol{X})$, are generated based on the sampled functions. These two augmented views are then fed into a shared encoder $f_\theta$, which is implemented using GCN \cite{GCN}, to extract the corresponding node representations: $\boldsymbol{Z}^1 = f_\theta(\boldsymbol{\widetilde{A}}^1, \boldsymbol{\widetilde{X}}^1)$ and $\boldsymbol{Z}^2 = f_\theta(\boldsymbol{\widetilde{A}}^2, \boldsymbol{\widetilde{X}}^2)$. We further $\ell_2$ normalize the node representations $\boldsymbol{Z}^1$ and $\boldsymbol{Z}^2$ to place them on the unit hypersphere, thereby the cross-view cosine similarity matrix can be computed as $\boldsymbol{S} = \boldsymbol{Z}^1 {\boldsymbol{Z}^2}^\top$. Finally, the model is optimized using a combination of self-alignment loss, node-neighbor alignment loss, and sparsity loss (refer to Section \ref{Sec: NSM Learning}).

\subsubsection{Node Clustering}
After training, we feed the original graph $\mathcal{G} = (\boldsymbol{A}, \boldsymbol{X})$ into the trained encoder $f_\theta$, resulting in node representations $\boldsymbol{Z} = f_\theta(\boldsymbol{A}, \boldsymbol{X})$. Following previous works \cite{EGAE, SCGC, CCGC}, we can apply the k-means algorithm directly to the $\ell_2$ normalized $\boldsymbol{Z}$ to obtain the clustering results.

\subsection{Node Similarity Matrix Learning}\label{Sec: NSM Learning}
We define node-wise similarity as a function that maps two arbitrary nodes to a semantic similarity indicator. A value of $1$ implies that the two nodes belong to the same ground-truth cluster, while a value of $0$ indicates nodes from distinct ground-truth clusters. Due to the absence of labels in the unsupervised clustering setting, we can not obtain the ideal node similarity matrix to guide representation learning. In this study, we try to approximate the ideal node similarity matrix by utilizing the cross-view consistency prior of contrastive learning, the homophily pattern in graphs, and the semantic-aware sparsification.

\subsubsection{Self Alignment}
One of the core components of contrastive learning is to enhance cross-view consistency, thereby learning essential and invariant representations across multiple views. This is based on the premise that different views of a given node are assuredly associated with the same semantic label. Technically, this consistency is achieved through the alignment of distinct augmented versions of the identical node in the representation space. In essence, it encourages the diagonal elements of the cross-view cosine similarity matrix $\boldsymbol{S}$ to converge towards $1$. So the self-alignment loss can be computed as:
\begin{equation}
    \mathcal{L}_{ali} = - \sum_{i=1}^n \boldsymbol{S}_{ii}.
    \label{Eq: Self Alignment}
\end{equation}

\subsubsection{Node-Neighbor Alignment}
From Table \ref{Tab: Dataset statistics}, Table \ref{Tab: Clustering Performance}, we can find that node homophily consistently exceeds the accuracy of clustering results for all datasets, thus providing empirical support for the homophily assumption, i.e., nodes exhibit a higher degree of semantic similarity with their neighbors. As a result, it is simple yet effective to align nodes and their neighbors in the representation space, which encourages the entries corresponding to node-neighbor pairs to converge towards $1$. Consequently, the computation of the node-neighbor loss is as follows:
\begin{equation}
    \mathcal{L}_{nei} = - \sum_{\boldsymbol{A}_{ij}=1} \boldsymbol{S}_{ij}. \label{Eq: Node-Neighbor Alignment}
\end{equation}

\subsubsection{Semantic-aware Sparsification}
We propose a semantic-aware sparsification method for the remaining disconnected node pairs, wherein edges are selectively added between nodes with relatively higher similarity. Specifically, we use the following differentiable equation to estimate the binary node similarity matrix:
\begin{equation}
    \boldsymbol{\hat{S}}_{ij} = \operatorname{Sigmoid} \left( \frac{\boldsymbol{S}_{ij}-s}{\tau} \right)=\frac{1}{e^{-\frac{\boldsymbol{S}_{ij}-s}{\tau}}+1}, \label{Eq: semantic-aware Sparsification}
\end{equation}
where $s \in [0, 1]$ represents the split cosine similarity score, and $\tau > 0$ is the temperature parameter. We opt for a relatively small temperature, such as $\tau = 0.1$, to ensure that $\boldsymbol{\hat{S}}_{ij}$ approaches values close to $1$ or $0$. 
Specifically, $\boldsymbol{\hat{S}}_{ij} \approx 1$ when $\boldsymbol{S}_{ij}$ exceeds $s$ by a little, and $\boldsymbol{\hat{S}}_{ij} \approx 0$ otherwise.  The differentiable node similarity matrix is expressed as $\boldsymbol{\hat{S}} = \operatorname{Sigmoid}\left(\frac{\boldsymbol{S} - s}{\tau}\right)$. Since the node similarity matrix is sparse (i.e., entries corresponding to inter-cluster nodes are $0$), we introduce a $\ell_1$ norm sparsity penalty to $\boldsymbol{\hat{S}}_{ij}$ for disconnected node pairs:
\begin{equation}
    \mathcal{L}_{spa} = \sum_{\boldsymbol{A}_{ij}=0}^{i \not= j} \parallel \boldsymbol{\hat{S}}_{ij} \parallel_1 = \sum_{\boldsymbol{A}_{ij}=0}^{i \not= j} \operatorname{Sigmoid} \left( \frac{\boldsymbol{S}_{ij}-s}{\tau} \right). \label{Eq: Sparsity Penalty}
\end{equation}

\begin{figure}
\centerline{\includegraphics[width=1.\linewidth]{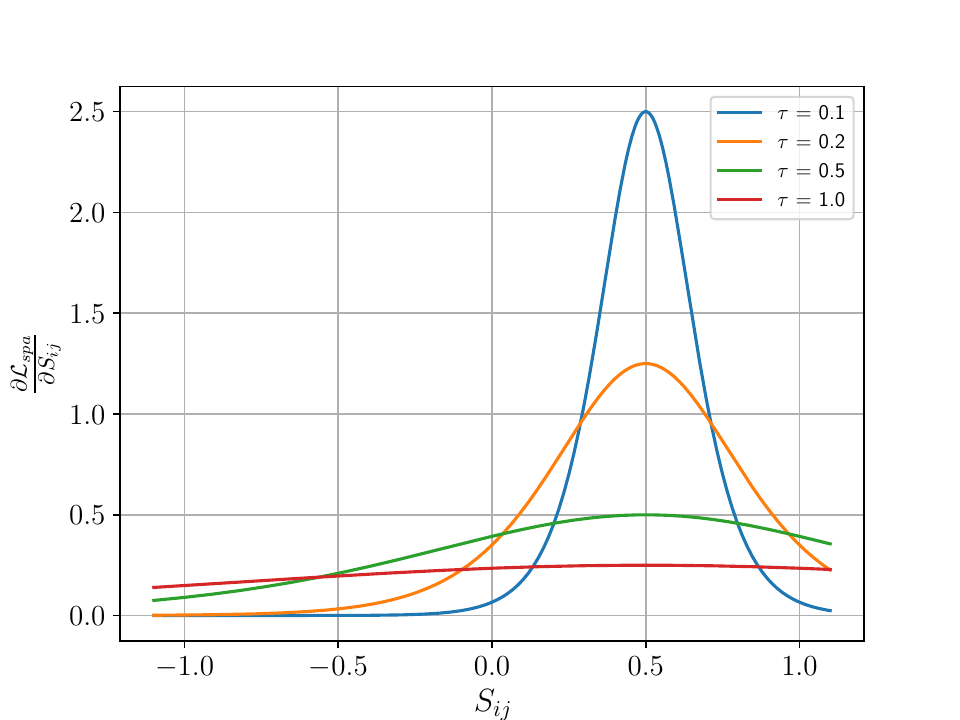}}
\caption{$\frac{\partial \mathcal{L}_{spa}}{\partial \boldsymbol{S}_{ij}}$ with $s = 0.5$ and different $\tau$ settings.} \label{Fig: Gradient Analysis}
\end{figure}

\textbf{Gradient Analysis.}\quad \label{Sec: Gradient Analysis}
Here we show that our sparsity penalty is adaptive and semantic-aware. The gradient of $\mathcal{L}_{spa}$ with respect to the pairwise cosine similarity $\boldsymbol{S}_{ij}$ is calculated as :
\begin{equation}
    \frac{\partial \mathcal{L}_{spa}}{\partial \boldsymbol{S}_{ij}}  = \frac{1/\tau}{e^{\frac{\boldsymbol{S}_{ij}-s}{\tau}}+2+e^{-\frac{\boldsymbol{S}_{ij}-s}{\tau}}}.
\end{equation}
The gradient magnitude can be viewed as the strength of the sparsity penalty. Figure \ref{Fig: Gradient Analysis} further illustrates a plot of gradient magnitudes $\frac{\partial \mathcal{L}_{spa}}{\partial \boldsymbol{S}_{ij}}$ with $s = 0.5$ and different $\tau$ settings over the domain $\boldsymbol{S}_{ij} \in [-1, 1]$. We have the following observations: 1) For node pairs more likely to be semantically different, i.e., $\boldsymbol{S}_{ij} < s$, the penalty becomes stronger as $\boldsymbol{S}_{ij}$ increases. Conversely, for more semantically similar node pairs, i.e., $\boldsymbol{S}_{ij} > s$, the penalty decreases with increasing $\boldsymbol{S}_{ij}$. Thus, the sparsity penalty's impact is notably lower on node pairs with very high semantic similarity (e.g., $\boldsymbol{S}_{ij} > 0.8$) or those already semantically different (e.g., $\boldsymbol{S}_{ij} < 0.2$), resulting in the optimization process being primarily influenced by node pairs around the split boundary. 2) As the temperature $\tau$ decreases, the distribution of the magnitude becomes more sharp. In other words, the sparsity penalty concentrates more on the split boundary region than the semantically similar region as the temperature decreases, and the sparsity penalty distribution tends to be more uniform as the temperature increases, which tends to give all node pairs the same magnitude of penalties. This is different from the InfoNCE \cite{InfoNCE} objective used in node-node contrastive methods \cite{GRACE, GCA, gCooL}, as InfoNCE penalizes more heavily for similar node pairs, potentially disrupting the underlying semantic structure \cite{UBCL}.

\subsubsection{Overall Objective}
The estimation of the node similarity matrix is expressed as $\boldsymbol{\hat{S}} = \operatorname{Sigmoid}\left(\frac{\boldsymbol{S} - s}{\tau}\right)$. Optimized by the self-alignment, the node-neighbor alignment, and the semantic-aware sparsification, the acquired node similarity matrix is expected to be reliable and appropriately sparse. Therefore, we formulate the overall objective as follows:
\begin{equation}
    \mathcal{L} = \mathcal{L}_{ali} + \lambda \mathcal{L}_{nei} + \gamma \mathcal{L}_{spa}, \label{Eq: Overall Objective}
\end{equation}
where $\lambda$ and $\gamma$ serve as weights that balance the contributions of three terms $\mathcal{L}_{ali}$, $\mathcal{L}_{nei}$ and $\mathcal{L}_{spa}$. Notably, we compute the mean, rather than the sum, of the elements within these terms in our implementation, which can result in nearly constant values for $\lambda$ and $\gamma$ across various datasets (as shown in Section \ref{Sec: Implementation Details}). To elucidate our method comprehensively, we present both a standard pseudocode in Algorithm \ref{ALG: NS4GC} and its PyTorch-style counterpart in Algorithm \ref{ALG: PyTorch-style pseudocode}.

\begin{algorithm}
\small
\caption{The overall procedure of NS4GC}
\label{ALG: NS4GC}
\flushleft{\textbf{Input}: Input graph $\mathcal{G}=(\boldsymbol{A},\boldsymbol{X})$; Hyperparameters $\lambda, \gamma, s, \tau, T, \mathcal{T}$.} \\
\flushleft{\textbf{Output}: The clustering result $\boldsymbol{C}$.} 
\begin{algorithmic}[1]
\FOR{$i=1$ to $T$}
\STATE Sample two augmentation functions $t^1 \sim \mathcal{T}$ and $t^2 \sim \mathcal{T}$;
\STATE Generate two augmented graphs via $(\boldsymbol{\widetilde{A}}^1, \boldsymbol{\widetilde{X}}^1) = t^1(\boldsymbol{A},\boldsymbol{X})$ and $(\boldsymbol{\widetilde{A}}^2, \boldsymbol{\widetilde{X}}^2)=t^2(\boldsymbol{A},\boldsymbol{X})$;
\STATE Obtain $\ell_2$ normalized node representations $\boldsymbol{Z}^1$ and $\boldsymbol{Z}^2$ using the same encoder $f_\theta$;
\STATE Compute the cross-view cosine similarity matrix $\boldsymbol{S}$ with $\boldsymbol{Z}^1 {\boldsymbol{Z}^2}^\top$;
\STATE Compute the self-alignment loss via Eq. (\ref{Eq: Self Alignment});
\STATE Compute the node-neighbor alignment loss via Eq. (\ref{Eq: Node-Neighbor Alignment});
\STATE Compute the semantic-aware sparsity loss via Eq. (\ref{Eq: Sparsity Penalty});
\STATE Update the parameters of $f_\theta$ by minimizing Eq. (\ref{Eq: Overall Objective}).
\ENDFOR
\STATE Perform k-means on $\ell_2$ normalized $\boldsymbol{Z}$ to obtain the final cluster assignments $\boldsymbol{C}$. 
\STATE \textbf{return} $\boldsymbol{C}$
\end{algorithmic}
\end{algorithm}

\begin{algorithm}[!ht]
   \caption{PyTorch-style pseudocode for NS4GC}
   \label{ALG: PyTorch-style pseudocode}
    \definecolor{codeblue}{rgb}{0.25,0.5,0.5}
    \lstset{
      basicstyle=\fontsize{8pt}{8pt}\ttfamily\bfseries,
      commentstyle=\fontsize{8pt}{8pt}\color{codeblue},
      keywordstyle=\fontsize{8pt}{8pt},
    }
\begin{lstlisting}[language=python]
# epochs
# s: split point
# tau: temperature
# lambda, gamma: trade-off

# X: node features
# A: adjacency matrix
# GNN: encoder $f_\theta$
# optimizer: Adam optimizer


# Get indices of connected nodes 
E = A._indices()
# Get masks of disconnected nodes
mask = torch.full_like(A, True)
mask[E] = False
mask.fill_diagonal_(False)  # excluding self-loops

# Pre-train
for _ in range(epochs): 
    # generate two views through random augmentation
    A1, X1 = augment(A, X)
    A2, X2 = augment(A, X)

    # obtain l2-normalized representations
    Z1 = normalize(GNN(A1, X1))
    Z2 = normalize(GNN(A2, X2))

    # compute the cosine similarity matrix
    S = Z1 @ Z2.T

    # compute self-alignment loss
    loss_ali = - torch.diag(S).mean()

    # compute node-neighbor alignment loss
    loss_nei = - S[E].mean()
    
    # compute semantic-aware sparsity loss 
    S = torch.masked_select(S, mask)
    loss_spa = torch.sigmoid((S - s) / tau).mean()

    loss = loss_ali + lambda*loss_nei + gamma*loss_spa
    
    # optimization step
    loss.backward()
    optimizer.step()

# Clustering
Z = GNN(A, X)
return KMeans(Z)
\end{lstlisting}
\end{algorithm}

\section{Experiments}
In this section, we design the experiments to evaluate our proposed NS4GC and answer the following research questions. \textbf{RQ1}: Does NS4GC outperform existing baseline methods on node clustering? \textbf{RQ2}: Can NS4GC learn more ideal node similarity matrix? \textbf{RQ3}: How does each loss component affect the clustering performance? \textbf{RQ4}: How do the parameters affect NS4GC?

\subsection{Experiment Setup}

\begin{table}
    \begin{center}
    {\caption{Dataset statistics.}\label{Tab: Dataset statistics}}
    \setlength{\tabcolsep}{2.5pt}
    \begin{tabular}{lcccccc}
    \toprule
    Dataset    & Type           & \#Nodes    & \#Edges   & \#Features  & \#Clusters  & Homo   \\
    \midrule 
    Cora       & citation       & 2,708      & 10,556    & 1,433       & 7           & 82.52\% \\
    Citeseer   & citation       & 3,327      & 9,228     & 3,703       & 6           & 72.22\% \\
    Pubmed     & citation       & 19,717     & 88,651    & 500         & 3           & 79.24\% \\
    CoraFull   & citation       & 19,793     & 126,842   & 8,710       & 70          & 58.61\% \\
    WikiCS     & reference      & 11,701     & 431,726   & 300         & 10          & 65.88\% \\
    Photo      & co-purchase    & 7,650      & 238,163   & 745         & 8           & 83.65\% \\
    Computer   & co-purchase    & 13,752     & 491,722   & 767         & 10          & 78.53\% \\
    CoauthorCS & co-authorship  & 18,333     & 163,788   & 6,805       & 15          & 83.20\% \\
    \bottomrule
    \end{tabular}
    \end{center}
\end{table}


\subsubsection{Datasets}
We evaluate our method on eight real-word datasets: Cora, Citeseer, Pubmed, CoraFull, WikiCS, Photo, Computer, and CoauthorCS. The detailed statistics are summarized in Table \ref{Tab: Dataset statistics}, and brief introductions are as follows:

\begin{itemize}
    \item \textbf{Cora}, \textbf{Citeseer}, \textbf{Pubmed} \cite{GCN} and \textbf{CoraFull} \cite{CoraFull} are four well-known citation network datasets, in which nodes represent publications and edges indicate their citations. All nodes are labeled based on the respective paper subjects.

    \item \textbf{WikiCS} \cite{WikiCS} is a reference network constructed from Wikipedia. It comprises nodes corresponding to articles in the field of Computer Science, where edges are derived from hyperlinks. The dataset includes 10 distinct classes representing various branches within the field. The node features are computed as the average GloVe word embeddings of the respective articles.

    \item \textbf{Photo} and \textbf{Computer} \cite{Amazon} are networks constructed from Amazon's co-purchase relationships. Nodes represent goods, and edges indicate frequent co-purchases between goods. The node features are represented by bag-of-words encoding of product reviews, and class labels are assigned based on the respective product categories.

    \item \textbf{CoauthorCS} \cite{Amazon} is a co-authorship network based on the Microsoft Academic Graph. Here, nodes are authors, that are connected by an edge if they co-authored a paper; node features represent paper keywords for each author’s papers, and class labels indicate the most active fields of study for each author.
\end{itemize}

\subsubsection{Baselines}
To verify the effectiveness of our proposed model, our evaluation includes a comprehensive comparison of NS4GC with fifteen baseline methods. These include one traditional clustering method, k-means; one community clustering method, DMoN \cite{DMoN}; one non-neural clustering method, SSGC \cite{SSGC}; four GAE-based clustering methods, VGAE \cite{VGAE}, DAEGC \cite{DAEGC}, GNAE \cite{GNAE}, and EGAE \cite{EGAE}; eight GCL-based clustering methods, DGI \cite{DGI}, MVGRL \cite{MVGRL}, GRACE \cite{GRACE}, gCooL \cite{gCooL}, CCASSG \cite{CCA-SSG}, BGRL \cite{BGRL}, SCGC \cite{SCGC}, and CCGC \cite{CCGC}.

\subsubsection{Evaluation Protocols}
We initiate the training by minimizing Eq. (\ref{Eq: Overall Objective}). The derived representations are then fed into a k-means model for node clustering \cite{EGAE, SCGC, CCGC}. Given the inherent randomness, we conduct the experiment twenty times across all comparison methods, reporting average values accompanied by their standard deviations. Four prevalent metrics, namely NMI, ARI, ACC, and F1 \cite{DAEGC, CCGC, SCGC}, are employed for evaluation.

\subsubsection{Implementation Details}\label{Sec: Implementation Details}
We employ the official implementations from GitHub for the baseline methods and develop NS4GC using PyTorch. The baseline results are either taken from the original paper or obtained using official codes when there are no corresponding dataset results in the original paper. 
All experiments are executed on an NVIDIA GeForce RTX 3090 GPU with 24 GB of memory. The graph encoder $f_\theta$ is specified as a standard two-layer GCN model \cite{GCN} for all datasets, with the exception of Citeseer, where empirical findings indicate that a one-layer GCN is more effective. The trade-off values, $\lambda$ and $\gamma$, are uniformly set to $1$. During training, we employ the Adam SGD optimizer \cite{Adam} with a learning rate and weight decay of $(1e-2, 1e-6)$ for the Computer dataset and $(1e-3, 1e-5)$ for the remaining datasets. Notably, we leverage the processed versions of all datasets as provided by the Deep Graph Library \cite{DGL}. And the number of clusters is pre-defined as the number of ground truth classes \cite{DAEGC, EGAE, SCGC, CCGC}. For a comprehensive overview of dataset-specific hyperparameters, please refer to Table \ref{Tab: Hypeparameter Specifications}.

\begin{table}
	\begin{center}
	\caption{Hypeparameter specifications.}
    \label{Tab: Hypeparameter Specifications}
    \begin{tabular}{lccccc}
    \toprule
    Dataset   & $p_{d1}, p_{d2}, p_{m1}, p_{m2}$  & $s$, $\tau$     & \#hid\_units  & \#epochs \\
    \midrule
    Cora      & 0.3, 0.3, 0.2, 0.2                & 0.6, 0.1        & 256-64    & 200    \\
    Citeseer  & 0.6, 0.8, 0.0, 0.4                & 0.5, 0.1        & 256       & 50     \\
    Pubmed    & 0.2, 0.6, 0.1, 0.0                & 0.6, 0.1        & 256-256   & 200    \\
    CoraFul   & 0.2, 0.4, 0.0, 0.2                & 0.6, 0.1        & 256-64    & 200    \\
    WikiCS    & 0.0, 0.6, 0.1, 0.0                & 0.5, 0.1        & 256-256   & 500    \\
    Photo     & 0.8, 0.8, 0.0, 0.0                & 0.6, 0.1        & 256-128   & 200    \\
    Computer  & 0.6, 0.8, 0.0, 0.1                & 0.6, 0.1        & 256-128   & 400    \\
    CoauthorCS& 0.1, 0.3, 0.0, 0.7                & 0.4, 0.08       & 256-64    & 200    \\
    \bottomrule
	\end{tabular}
    \end{center}
\end{table}

\begin{table*}[!ht]
    \centering
    \small
    \caption{Overall performance of graph clustering on eight datasets measured by ARI, NMI, ACC, and F1 scores in percentage. }\label{Tab: Clustering Performance}
    \renewcommand\arraystretch{1.1}
    \resizebox{2\columnwidth}{!}{
          \begin{tabular}{c|c|c|c|c|c|c|c|c|c|c}
            \bottomrule
                                \textbf{Method}       & \textbf{Metric} & \textbf{Cora}       & \textbf{Citeseer}   & \textbf{Pubmed}     & \textbf{CoraFull}        & \textbf{WikiCS}       & \textbf{Photo}     & \textbf{Computer} & \textbf{CoauthorCS}   & \textit{Rank}  \\ 
            \hline
            \multirow{3}{*}{\textbf{k-means}} & NMI             & 15.44±3.83 & 20.66±2.83 & 31.34±0.15 & 34.52±0.76 & 25.16±0.31 & 32.61±0.38 & 24.26±0.45 & 66.36±0.70    & 14.8 \\   
                                              & ARI             & 9.49±2.01 & 16.80±3.02 & 28.12±0.03 & 9.03±0.62 & 14.50±0.34 & 20.66±0.91   & 9.36±0.42  & 54.47±2.28    & 14.8  \\   
                                              & ACC             & 34.07±3.21 & 43.78±3.60 & 60.11±0.03 & 25.61±1.12 & 33.90±1.09 & 41.52±0.71 & 26.30±0.48 & 60.13±2.64    & 15.9 \\ 
                                              & F1              & 31.39±4.29 & 41.43±3.72 & 59.23±0.08 & 11.39±0.58 & 30.17±1.30 & 40.92±1.61 & 21.16±0.37 & 52.26±2.84    & 15.9 \\ \hline
           \multirow{3}{*}{\textbf{DMoN}}     & NMI             & 49.37±0.57 & 32.19±0.39 & 29.33±0.67 & 52.57±0.68 & 40.14±0.76 & 62.09±0.45 & 48.32±0.78 & 70.59±0.24    & 13.5    \\   
                                              & ARI             & 38.91±0.54 & 32.47±0.41 & 27.30±0.51 & 24.16±0.63 & 29.90±0.62 & 51.77±0.52 & 26.92±0.81 & 55.59±0.39    & 13.1   \\   
                                              & ACC             & 62.64±0.48 & 57.59±0.67 & 66.42±0.46 & 35.25±0.96 & 44.54±0.82 & 70.04±0.63 & 37.52±0.71 & 63.74±0.54    & 13.4  \\ 
                                              & F1              & 57.67±0.45 & 53.53±0.59 & 62.69±0.33 & 28.25±0.92 & 36.37±0.78 & 59.87±0.66 & 26.85±0.90 & 53.87±0.65    & 14.4  \\ \hline
            \multirow{3}{*}{\textbf{SSGC}}    & NMI             & 54.32±1.92 & 42.97±0.08 & 32.27±0.01 & \underline{56.41±0.18} & 43.67±0.06 & 66.44±0.96 & 52.05±0.73 & 72.19±0.41  & 6.4   \\   
                                              & ARI             & 46.27±4.01 & 43.54±0.15 & 31.06±0.01 & \underline{29.89±0.89} & 33.35±0.07 & 51.65±2.40 & 38.29±1.03 & 61.58±1.00  & 7.1 \\   
                                              & ACC             & 69.28±3.70 & 68.23±0.18 & 68.61±0.01 & 41.46±1.25 & \underline{54.81±0.07} & 68.71±3.06 & \underline{55.21±0.49} & 69.68±1.23  & 6.3 \\ 
                                              & F1              & 64.70±5.53 & 64.16±0.13 & 68.22±0.01 & \underline{34.57±1.10} & 44.88±0.10 & 61.96±3.52 & 42.49±2.73 & 64.99±2.19 & 6.3 \\ \hline
            \multirow{3}{*}{\textbf{VGAE}}    & NMI             & 52.48±1.33 & 34.46±0.92 & 27.16±1.45 & 49.17±0.17 & 43.64±0.75 & 67.36±1.58 & 50.50±1.26 & 68.49±1.11   & 11.4   \\   
                                              & ARI             & 43.99±2.34 & 32.65±0.92 & 26.32±1.15 & 19.80±0.44 & 32.97±0.90 & 57.00±2.55 & 30.61±2.16 & 50.64±3.12   & 12.9 \\   
                                              & ACC             & 64.54±1.98 & 59.97±0.72 & 66.08±0.71 & 31.23±0.58 & 49.69±0.66 & 73.66±2.27 & 46.08±1.81 & 61.80±1.69   & 13.0  \\ 
                                              & F1              & 64.50±1.37 & 57.29±0.59 & 64.85±0.80 & 25.81±0.54 & 42.03±0.67 & 69.94±1.87 & 40.31±4.01 & 53.67±1.15   & 12.4 \\ \hline   
            \multirow{3}{*}{\textbf{DAEGC}}   & NMI             & 52.89±0.69 & 39.41±0.86 & 28.26±0.03 & 49.16±0.73 & 41.89±0.60 & 65.57±0.03 & 50.87±1.45 & 70.52±1.23    & 11.5  \\   
                                              & ARI             & 49.63±0.43 & 40.78±1.24 & 29.84±0.04 & 22.60±0.47 & 33.24±1.50 & \underline{59.39±0.02} & 31.82±2.34 & 54.23±2.94 & 9.5   \\   
                                              & ACC             & 70.43±0.36 & 67.54±1.39 & 68.73±0.03 & 34.35±1.00 & 51.92±1.06 & 76.44±0.01 & 49.22±1.56 & 64.22±1.88   & 9.0 \\ 
                                              & F1              & 68.27±0.57 & 62.20±1.32 & 68.23±0.02 & 26.96±1.33 & 42.09±0.71 & 69.97±0.02 & 42.45±3.43 & 59.02±2.42   & 9.1  \\ \hline
            \multirow{3}{*}{\textbf{GNAE}}    & NMI             & 54.20±0.60 & 33.93±0.02 & 30.56±0.17 & 54.42±0.23 & 43.33±0.50 & \underline{68.96±0.00} & 50.22±0.75 & 73.81±0.68  & 8.1  \\   
                                              & ARI             & 47.74±1.66 & 33.00±0.02 & 31.00±0.15 & 27.14±0.83 & 34.69±0.48 & 58.83±0.00 & 34.45±0.65 & 57.71±1.13  & 7.9  \\   
                                              & ACC             & 69.12±1.50 & 58.96±0.03 & 69.03±0.09 & 39.43±1.12 & 52.30±0.33 & 77.10±0.00 & 53.74±0.63 & 65.62±2.36  & 7.4  \\ 
                                              & F1              & 68.67±1.13 & 54.96±0.03 & 67.94±0.11 & 30.82±1.07 & 43.58±0.41 & 70.81±0.00 & 42.97±0.77 & 61.14±4.13  & 7.3  \\ \hline
            \multirow{3}{*}{\textbf{EGAE}}    & NMI             & 52.83±0.78 & 38.69±1.76 & 32.24±0.18 & 53.92±1.01 & 42.36±1.05 & 61.77±1.42 & 52.25±0.62 & 71.18±0.98   & 9.9  \\   
                                              & ARI             & 50.32±0.73 & 40.51±2.76 & \underline{32.67±0.09} & 27.18±0.76 & 30.26±1.02 & 50.43±8.35 & 36.69±0.71 & 53.98±0.82 & 8.9  \\   
                                              & ACC             & 72.86±0.46 & 66.21±1.89 & 68.98±0.21 & 39.97±0.87 & 48.73±1.67 & 73.67±0.79 & 53.12±0.35 & 63.67±1.37  & 8.3  \\ 
                                              & F1              & 68.67±0.58 & 61.03±2.32 & 67.69±0.14 & 28.52±0.77 & 41.92±1.04 & 69.30±0.76 & 41.52±0.46 & 60.80±1.26 & 9.5   \\ \hline
            \multirow{3}{*}{\textbf{DGI}}     & NMI             & 55.82±0.60 & 41.16±0.54 & 25.27±0.02 & 54.61±0.27 & 43.75±0.25 & 64.79±0.42 & 51.30±0.86 & 75.18±0.70   & 7.8  \\   
                                              & ARI             & 48.91±1.42 & 39.78±0.74 & 24.06±0.03 & 25.26±0.43 & 33.64±0.63 & 56.19±0.70 & 39.70±1.09 & 63.39±1.99   & 8.6 \\   
                                              & ACC             & 70.89±1.05 & 63.31±1.20 & 65.01±0.02 & 36.52±0.56 & 51.54±0.82 & 75.78±0.67 & 54.36±1.20 & 70.77±2.56   & 8.1  \\ 
                                              & F1              & 68.37±1.44 & 60.00±1.83 & 65.38±0.02 & 31.58±0.63 & 44.13±1.71 & 70.52±1.16 & 39.95±1.52 & \underline{68.87±3.86}   & 7.8 \\ \hline
            \multirow{3}{*}{\textbf{MVGRL}}  &  NMI             & 56.30±0.27 & 43.47±0.08 & 27.07±0.00 & 52.81±0.22 & 39.22±0.13 & 62.16±0.71 & 50.07±0.51 & 75.15±0.61   & 10.0  \\   
                                              & ARI             & 50.28±0.40 & 44.09±0.09 & 24.53±0.00 & 23.47±0.54 & 28.08±0.17 & 48.71±1.64 & 37.10±2.05 & 63.26±1.03   & 10.1 \\   
                                              & ACC             & 72.03±0.17 & 68.03±0.06 & 64.11±0.00 & 35.18±0.84 & 45.26±0.18 & 70.54±0.66 & 53.22±1.13 & 69.15±2.19   & 9.9  \\ 
                                              & F1              & 68.36±0.53 & 63.66±0.04 & 64.96±0.00 & 29.92±0.74 & 40.28±0.15 & 64.58±0.93 & 38.38±0.40 & 64.02±3.52   & 10.1   \\ \hline
            \multirow{3}{*}{\textbf{GRACE}}   & NMI             & 51.98±0.24 & 39.07±0.07 & 22.44±0.02 & 54.27±0.18 & 44.69±0.23 & 66.62±0.02 & 52.33±0.91 & 72.58±0.26   & 8.9  \\   
                                              & ARI             & 46.20±0.53 & 40.38±0.08 & 15.62±0.01 & 25.90±0.53 & 32.72±0.70 & 57.01±0.07 & 34.18±2.59 & 55.62±0.47   & 10.8  \\ 
                                              & ACC             & 68.28±0.30 & 65.83±0.06 & 56.30±0.01 & 37.38±0.82 & 50.82±1.99 & 74.87±0.05 & 50.62±1.97 & 63.64±1.48   & 11.3 \\ 
                                              & F1              & 65.67±0.44 & 62.59±0.05 & 56.40±0.01 & 32.41±0.76 & 43.31±2.25 & 70.58±0.03 & 41.14±2.50 & 60.23±2.76   & 9.3 \\ \hline
            \multirow{3}{*}{\textbf{gCooL}}   & NMI             & 50.25±1.08 & 41.67±0.41 & \textbf{33.14±0.02} & 54.21±0.39 & \underline{46.94±0.03} & 65.32±0.15 & 45.67±0.84 & 73.90±1.13  & 7.8  \\   
                                              & ARI             & 44.95±1.74 & 42.66±0.47 & 31.93±0.01 & 25.52±0.68 & \underline{38.33±0.02} & 56.98±0.13 & 38.62±0.36 & 59.43±3.15  & 6.8 \\
                                              & ACC             & 67.40±1.43 & 66.98±0.32 & \underline{69.97±0.02} & 38.93±0.96 & 53.64±0.23 & 75.97±0.01 & 50.09±1.10 & 66.19±3.14  & 7.4 \\  
                                              & F1              & 65.50±1.17 & 63.23±0.33 & \underline{68.90±0.03} & 32.42±1.01 & \underline{45.04±0.07} & \underline{73.10±0.14} & 30.10±1.07 & 61.00±4.17 & 6.5 \\ \hline
            \multirow{3}{*}{\textbf{CCASSG}}  & NMI             & \underline{56.51±1.49} & 43.69±0.24 & 29.61±0.01 & 55.17±0.19 & 45.35±0.07 & 63.89±0.02 & 52.32±0.41 & 72.43±0.42   & 6.1  \\   
                                              & ARI             & 50.77±3.39 & 44.26±0.23 & 25.81±0.01 & 27.37±0.57 & 36.65±0.04 & 54.70±0.01 & \underline{40.88±0.24} & 59.01±1.94   & 6.3  \\   
                                              & ACC             & 71.89±2.52 & 69.26±0.20 & 64.49±0.01 & 38.01±0.59 & 53.84±0.03 & 74.03±0.02 & 53.43±0.03 & 65.83±1.15 & 7.5     \\ 
                                              & F1              & \underline{70.98±1.65} & 63.87±0.21 & 63.96±0.01 & 31.42±0.63 & 43.19±0.02 & 67.74±0.02 & \underline{42.99±0.46} & 57.18±1.76  & 7.8 \\ \hline
            \multirow{3}{*}{\textbf{BGRL}}    & NMI             & 55.46±0.81 & 41.43±1.26 & 30.05±0.02 & 54.40±0.37 & 46.54±0.40 & 67.13±0.71 & \underline{54.87±0.26} & \underline{76.31±0.54}  & \underline{5.4} \\   
                                              & ARI             & 50.75±1.68 & 38.70±1.90 & 27.32±0.01 & 24.13±0.65 & 37.44±1.35 & 57.59±0.65 & 34.62±1.51 & \underline{66.04±1.53}   & 7.3  \\   
                                              & ACC             & 71.69±1.59 & 65.59±1.17 & 65.84±0.00 & 38.02±0.89 & 53.47±1.55 & 75.67±0.59 & 49.30±1.28 & \underline{72.70±1.81}   & 8.0  \\ 
                                              & F1              & 69.87±1.67 & 59.67±2.15 & 65.61±0.00 & 33.52±0.84 & 44.79±1.34 & 70.55±0.36 & 42.72±1.05 & 67.40±3.17 & \underline{6.0}   \\ \hline  
            \multirow{3}{*}{\textbf{SCGC}}    & NMI             & 56.10±0.72 & \underline{45.25±0.45} & 31.17±0.81 & 53.00±0.29 & 42.74±1.07 & 67.67±0.88 & 51.01±0.89 & 70.21±0.80    & 7.5 \\   
                                              & ARI             & 51.79±1.59 & \textbf{46.29±1.13} & 29.23±0.55 & 24.23±0.86 & 35.96±1.10 & 58.48±0.72 & 34.39±0.73 & 60.43±0.55 & 6.1 \\   
                                              & ACC             & \underline{73.88±0.88} & \textbf{71.02±0.77} & 68.66±0.45 & \underline{41.89±0.47} & 50.53±1.03 & \underline{77.48±0.37} & 51.76±1.05 & 69.18±0.55  & 4.9  \\ 
                                              & F1              & 70.81±1.96 & \underline{64.80±1.01} & 68.05±0.31 & 32.98±0.73 & 40.34±0.95 & 72.22±0.97 & 39.03±1.65 & 62.76±1.21 & 6.4  \\   \hline    
            \multirow{3}{*}{\textbf{CCGC}}    & NMI             & 56.45±1.04 & 44.33±0.79 & 32.10±0.23 & 52.88±0.23 & 42.24±1.37 & 67.44±0.48 & 53.79±0.59 & 73.51±0.56   & 6.0   \\   
                                              & ARI             & \underline{52.51±1.89} & 45.68±1.80 & 30.83±0.41 & 27.02±0.49 & 35.64±0.03 & 54.78±0.04 & 39.84±0.75 & 63.93±0.35   & \underline{4.9} \\   
                                              & ACC             & 73.88±1.20 & \underline{69.84±0.94} & 68.78±0.56 & 39.69±0.49 & 49.61±1.49 & 77.25±0.41 & 54.70±0.81 & 70.16±0.88   & \underline{4.6} \\ 
                                              & F1              & 70.98±2.79 & 62.71±2.06 & 68.35±0.71 & 26.98±0.73 & 41.12±1.26 & 72.18±0.57 & 42.09±0.74 & 67.63±0.71 & 6.5   \\ \hline                 
            \multirow{3}{*}{\textbf{NS4GC}}   & NMI             & \textbf{60.34±0.09} & \textbf{45.37±0.08} & \underline{32.76±0.01} & \textbf{57.09±0.17} & \textbf{48.23±0.32} & \textbf{72.55±0.02} & \textbf{57.29±0.00} & \textbf{78.63±0.90}  & \textbf{1.1}   \\   
                                              & ARI             & \textbf{58.00±0.46} & \underline{46.08±0.08} & \textbf{33.50±0.01} & \textbf{30.42±0.65} & \textbf{40.39±1.40} & \textbf{62.58±0.02} & \textbf{44.56±0.00} & \textbf{75.92±3.68}   & \textbf{1.1} \\   
                                              & ACC             & \textbf{76.33±0.15} & 69.70±0.03 & \textbf{70.56±0.01} & \textbf{42.12±0.56} & \textbf{57.39±1.89} & \textbf{79.38±0.01} & \textbf{58.21±0.00} & \textbf{78.63±2.22}  & \textbf{1.3}  \\ 
                                              & F1              & \textbf{74.23±0.08} & \textbf{65.08±0.08} & \textbf{69.49±0.01} & \textbf{34.84±0.63} & \textbf{49.98±2.35} & \textbf{73.33±0.01} & \textbf{48.58±0.00} & \textbf{73.68±3.83}  & \textbf{1.0} \\
            \toprule
            \end{tabular}
    }
\end{table*}

\subsection{Clustering Performance (\textbf{RQ1})}
The clustering performance of all compared methods across eight datasets is presented in Table \ref{Tab: Clustering Performance}, where bold and underlined values indicate the best and the runner-up results, respectively. Our proposed NS4GC achieves state-of-the-art performance, substantially improving the performance benchmark set by contrastive graph clustering methods. For instance, on the CoauthorCS dataset, our method outperforms the runner-up by 2.32\%, 9.88\%, 5.93\%, and 4.81\% with respect to NMI, ARI, ACC, and F1. The superior performance of our method can be attributed to two factors: 1) the contrastive mechanism enables the network to capture more supervision information from multiple views, leading GCL-based clustering methods to generally outperform GAE-based methods. 2) Its cutting-edge objective, aiming to learn the approximately ideal node similarity matrix, ensures representations of semantically similar nodes are positioned closely within the representation space, resulting in clustering-friendly representations.

\begin{table*}
    \begin{center}
    \renewcommand{\arraystretch}{1.2}
    {\caption{ Ablation study of node clustering performance on the key components of NS4GC (in percentage).}\label{Tab: Ablation study}}
    \begin{tabular}{ccc|cc|cc|cc|cc}
    \bottomrule
    \multirow{2}{*}{$\mathcal{L}_{ali}$} & \multirow{2}{*}{$\mathcal{L}_{nei}$} & \multirow{2}{*}{$\mathcal{L}_{spa}$} & \multicolumn{2}{c|}{Cora}      & \multicolumn{2}{c|}{WikiCS}      & \multicolumn{2}{c|}{Photo}   & \multicolumn{2}{c}{CoauthorCS} \\
    \cline{4-11}
    ~ & ~ & ~ & ACC      & NMI        & ACC        & NMI        & ACC        & NMI        & ACC        & NMI     \\ \hline
    \checkmark & \checkmark & ~ & 53.73±2.51 & 35.92±1.31 & 53.32±3.45 & 42.66±2.12 & 65.46±5.23 & 59.74±2.77 & 65.14±1.88 & 71.84±0.53 \\
    \checkmark & ~ & \checkmark & 71.60±3.36 & 55.87±2.85 & 48.12±2.36 & 44.70±0.68 & 71.81±1.95 & 68.99±0.83 & 75.09±1.00 & 76.72±0.41 \\
    ~ & \checkmark & \checkmark & 74.83±2.22 & 58.33±1.37 & 51.18±0.45 & 45.60±0.26 & 76.06±0.01 & 66.90±0.01 & 65.77±1.97 & 71.60±0.42 \\
    \checkmark & \checkmark & \checkmark & 76.33±0.15 & 60.34±0.09 & 57.39±1.89 & 48.23±0.32 & 79.38±0.01 & 72.55±0.02 & 78.63±2.22 & 78.63±0.90 \\
    \toprule
    \end{tabular}
    \end{center}
\end{table*}

It is noteworthy that both GRACE and CCASSG adopt the same network architecture as NS4GC, differing primarily in the constraints imposed on the node similarity matrix within the representation space. Specifically, GRACE implicitly assumes the node similarity matrix to be an identity matrix, CCASSG does not exploit the sparsity of this matrix, while NS4GC incorporates a novel semantic-aware sparsification. The superior performance of NS4GC substantiates the efficacy of this proposed semantic-aware sparsification. Moreover, in comparison to gCooL, which integrates community/cluster detection for learning cluster-wise similarity, and CCGC, which incorporates a time-consuming k-means at each epoch to construct true positive samples and hard negative samples by mining high-confidence clustering information, NS4GC also demonstrates superior performance, providing empirical evidence of the effectiveness of simple node-neighbor alignment for homophilic graphs.

\begin{figure}[!ht]
    \centering  
    \subfigure[GRACE]{\includegraphics[width=0.49\linewidth]{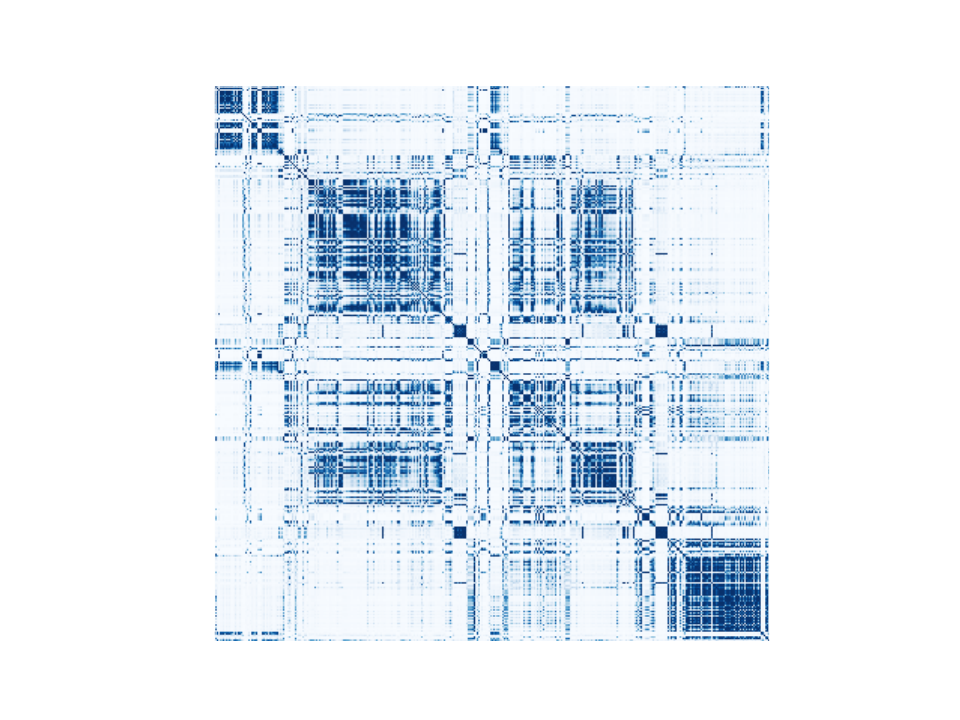}}
    \subfigure[CCASSG]{\includegraphics[width=0.49\linewidth]{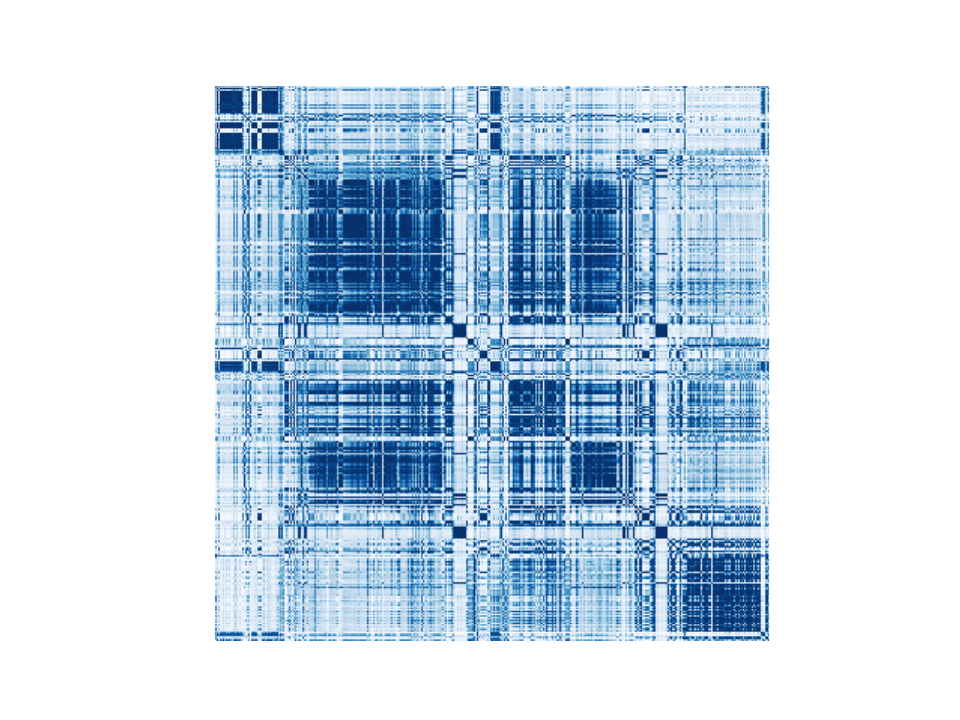}}
    \subfigure[NS4GC]{\includegraphics[width=0.49\linewidth]{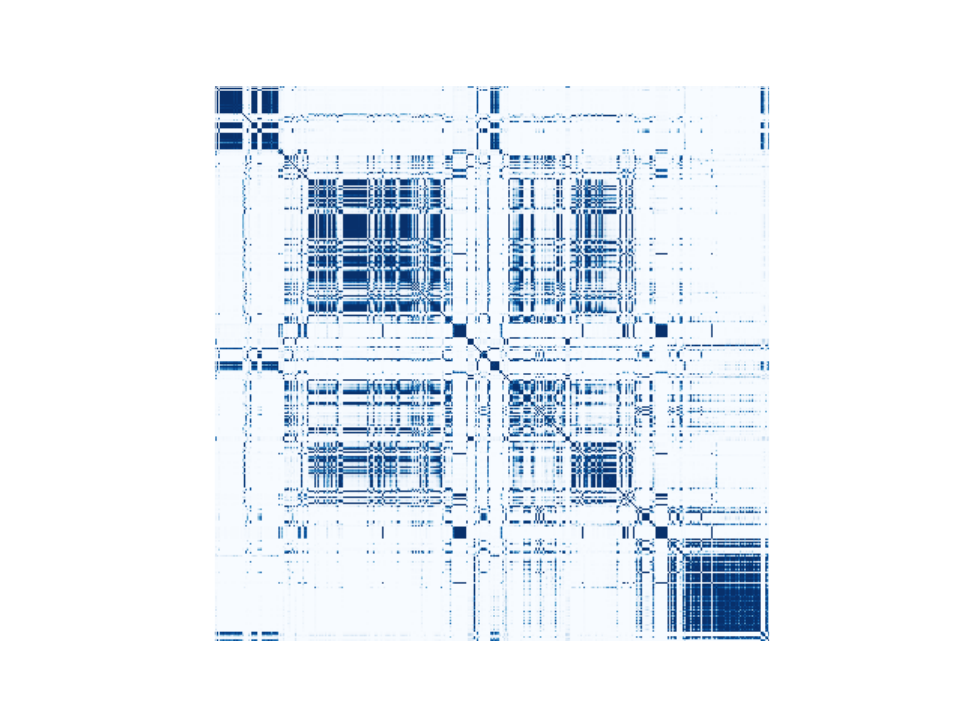}}
    \subfigure[Ideal]{\includegraphics[width=0.49\linewidth]{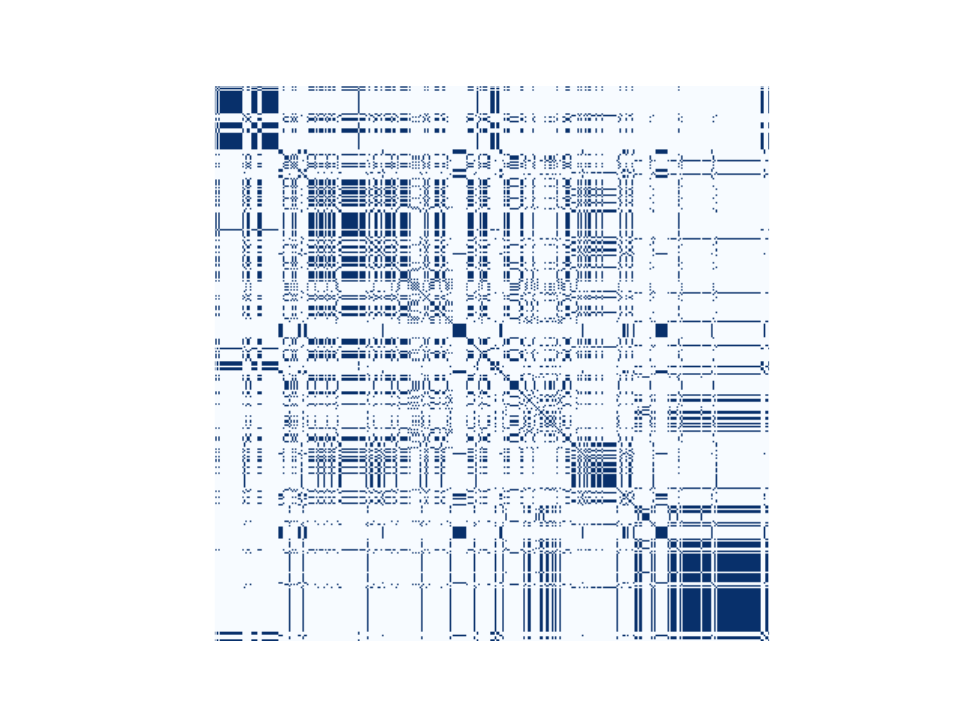}}
    \caption{Node similarity matrices learned by GRACE, CCASSG, and NS4GC and the ideal node similarity matrix on Photo.}
    \label{Fig: NSM Visualization}
\end{figure}

\subsection{Visualization of Node Similarity Matrix (\textbf{RQ2})}
To gain visual insights into our method, we present visualizations illustrating the cosine similarity matrix (following the sparsification defined in Eq. (\ref{Eq: semantic-aware Sparsification}) with $s=0.6$ and $\tau=0.1$) of the representations learned by GRACE, CCASSG, and NS4GC using the Photo dataset. As depicted in Figure \ref{Fig: NSM Visualization}, several observations emerge: 
1) A conspicuous block-diagonal structure is evident in these node similarity matrices.
2) The off-diagonal elements of CCASSG's node similarity matrix show a significant increase compared to GRACE and NS4GC, due to its absence of explicit constraints on off-diagonal elements.
3) While the node similarity matrices of GRACE and our method share a similar sparse structure, our diagonal blocks exhibit a deeper color, and non-diagonal-block areas appear slightly lighter, demonstrating the role of the proposed node-neighbor alignment and a semantic-aware sparsification.
4) The node similarity matrix learned by NS4GC exhibits higher similarity to the ideal node similarity matrix.

\begin{table}[!ht]
  \begin{center}
  \setlength{\tabcolsep}{4pt}
  {\caption{Comparison between the learned node similarity matrices and the ideal node similarity matrix (in percentage).}}\label{Tab: Comparison of Node Similarity Matrix}
  \renewcommand\arraystretch{1.2}
  \begin{tabular}{l|cc|cc|cc|cc}
  \bottomrule
  Dataset    & \multicolumn{2}{c|}{Cora}  & \multicolumn{2}{c|}{WikiCS}  & \multicolumn{2}{c|}{Photo}  & \multicolumn{2}{c}{CoauthorCS} \\ \hline
  Metric   & MAE        & ACC        & MAE        & ACC        & MAE        & ACC        & MAE        & ACC        \\ \hline
    DGI & 19.74 & 84.13 & 15.38 & 86.54 & 16.03 & 87.92 & 7.96 & 94.71 \\
    GRACE & 21.55 & 82.92 & 14.08 & 86.66 & 11.78 & 89.91 & 8.35 & 93.02 \\
    CCASSG & 17.18 & 84.35 & 15.66 & 86.41 & 15.88 & 88.75 & 8.90 & 92.66 \\
    BGRL & 20.90 & 83.61 & 15.62 & 85.22 & 18.68 & 82.57 & 13.20 & 88.66 \\
    CCGC & 45.87 & 82.11 & 18.12 & 84.47 & 37.03 & 86.77 & 18.97 & 91.51 \\
    NS4GC & \textbf{15.65} & \textbf{85.26} & \textbf{13.83} & \textbf{87.17} & \textbf{11.12} & \textbf{90.53} & \textbf{4.76} & \textbf{96.28} \\
  \toprule
  \end{tabular}
  \end{center}
\end{table}

\subsection{Comparison of Node Similarity Matrix (\textbf{RQ2})}
Here we further quantitatively assess disparities between the node similarity matrices obtained from various contrastive graph clustering methods and the ideal node similarity matrix. Specifically, considering the one-hot ground-truth cluster labels $\boldsymbol{C}$ assigned to all nodes, we compute the ideal node similarity matrix as $\boldsymbol{N} = \boldsymbol{C}\boldsymbol{C}^\top$. 
Using the cosine similarity matrix $\boldsymbol{S}$ derived from the learned node representations $\boldsymbol{Z}$, we softly estimate the learned node similarity matrix $\boldsymbol{W}$ with Eq. (\ref{Eq: semantic-aware Sparsification}), i.e., $\boldsymbol{W}=\operatorname{Sigmoid} \left( (\boldsymbol{S}-s)/0.1 \right)$. To gauge the dissimilarity, we employ the mean absolute error (MAE) between $\boldsymbol{W}$ and $\boldsymbol{N}$, i.e., $\operatorname{MAE} = \frac{1}{n^2}\sum_i \sum_j |\boldsymbol{M}_{ij}-\boldsymbol{N}_{ij}|$.
Additionally, we explore binarizing the cosine similarity matrix $\boldsymbol{S}$ to form the learned node similarity matrix, expressed as $\boldsymbol{W}=\mathbb{I}(\boldsymbol{S} \geq s)$, where $\mathbb{I}(\cdot)$ denotes the indicator function. And we use the accuracy as our metric, i.e., $\operatorname{ACC} = \frac{1}{n^2}\sum_i\sum_j \mathbb{I}(\boldsymbol{W}_{ij}=\boldsymbol{N}_{ij})$. To ensure fairness in comparison, we search for the optimal split value $s$ from the set $\{ 0.6, 0.7, 0.8, 0.9, 0.95 \}$ for each method and dataset. The results presented in Table \ref{Tab: Comparison of Node Similarity Matrix} demonstrate the consistent learning of the most similar node similarity matrix to the ideal node similarity matrix by our proposed method.

\subsection{Ablation Study (\textbf{RQ3})}
To systematically assess the impact of each loss component, we conduct ablation studies on the Cora, WikiCS, Photo, and CoauthorCS datasets, involving different combinations of the self-alignment $\mathcal{L}_{ali}$, the node-neighbor alignment $\mathcal{L}_{nei}$, and the semantic-aware sparsification $\mathcal{L}_{spa}$. The results are collated in Table \ref{Tab: Ablation study}, revealing the following key observations: 1) Solely utilizing the self-alignment loss and the node-neighbor alignment loss resulted in all node representations converging towards uniformity, leading to suboptimal performance. 2) The introduction of the semantic-aware sparsification term had a notable positive impact on performance by alleviating representation collapse. 3) By integrating all these loss terms, our model is guided by an approximately ideal node similarity matrix, yielding significant performance enhancements. For example, on the Cora dataset, NS4GC improves ACC by about 23\% , and NMI by about 25\%. These results highlight the effectiveness of incorporating the node-neighbor alignment and the semantic-aware sparsification to learn clustering-friendly node representations.

\begin{figure}[!ht]
    \centering
    \subfigure[Cora]{\includegraphics[width=0.49\linewidth]{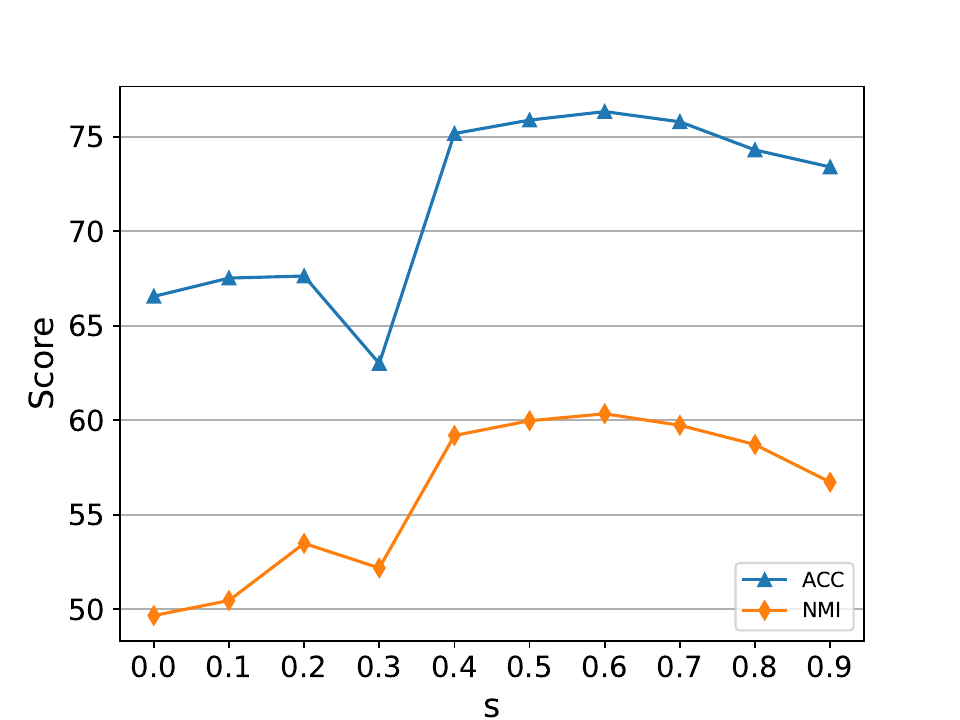}}  
    \subfigure[WikiCS]{\includegraphics[width=0.49\linewidth]{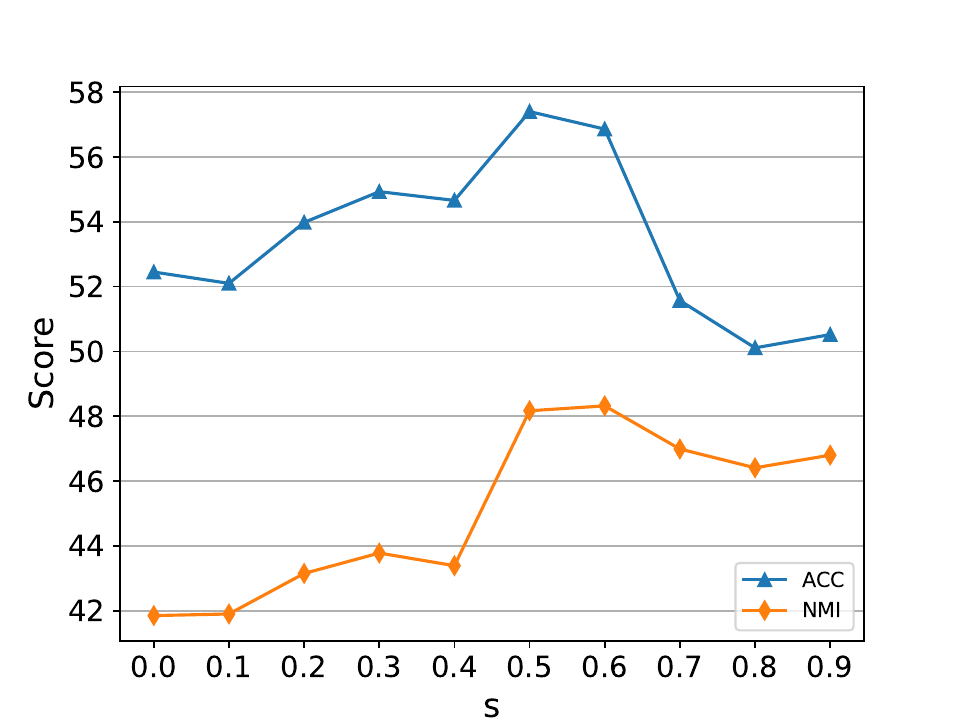}}
    \subfigure[Photo]{\includegraphics[width=0.49\linewidth]{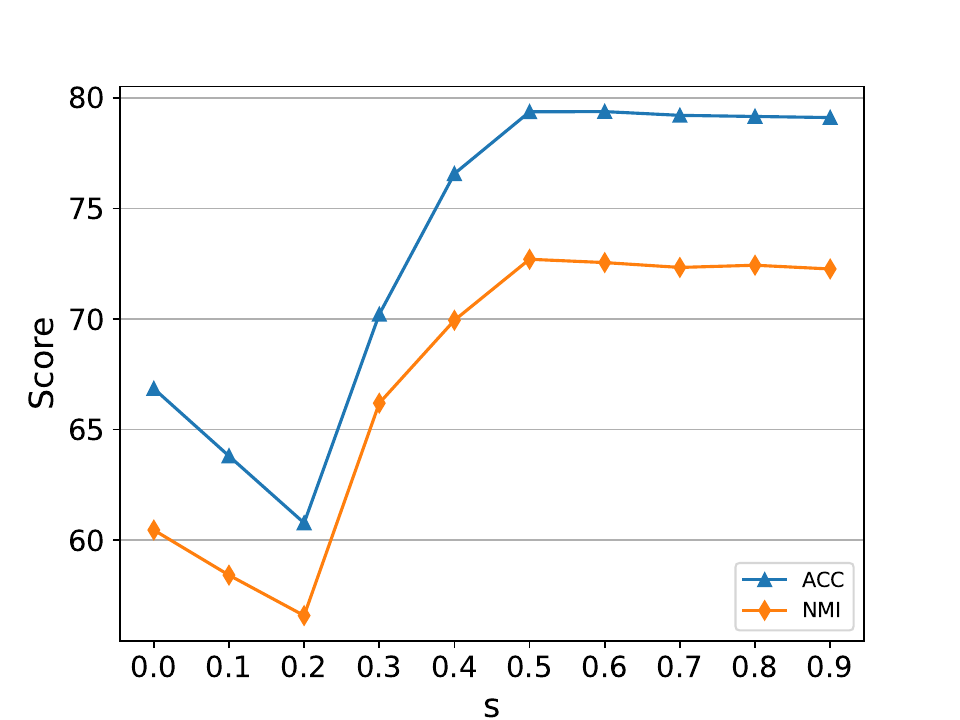}}
    \subfigure[CoauthorCS]{\includegraphics[width=0.49\linewidth]{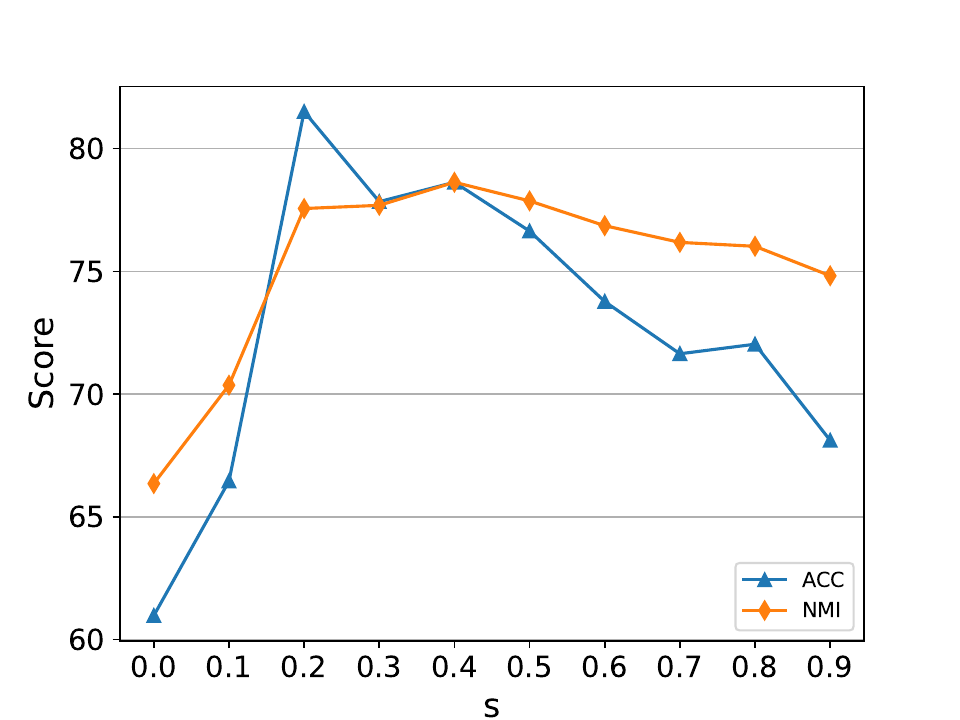}}
    \caption{Impact of the split value $s$ on Cora, WikiCS, Photo and CoauthorCS.}
    \label{Fig: param_s}
\end{figure}



\subsection{Hyperparameter Analysis (\textbf{RQ4})}
In this subsection, we analyze the sensitivity of NS4GC with respect to the hyperparameters: the split value $s$, the temperature $\tau$, the trade-off $(\lambda, \gamma)$, and the augmentation intensity $(p_{d}, p_m)$.

\textbf{Impact of the split value $s$.} We investigate the impact of the split cosine similarity score $s$ on NS4SC. Figure \ref{Fig: param_s} illustrates the ACC and NMI metrics for four datasets across a range of $s$ values, spanning from 0.0 to 0.9. Since too small/large $s$ would lead to an under-sparse/over-sparse node similarity matrix within the representation space, we recommend selecting a split value in the vicinity of 0.5, such as 0.6, for new datasets.

\begin{figure}[!ht]
    \centering
    \subfigure[Cora]{\includegraphics[width=0.49\linewidth]{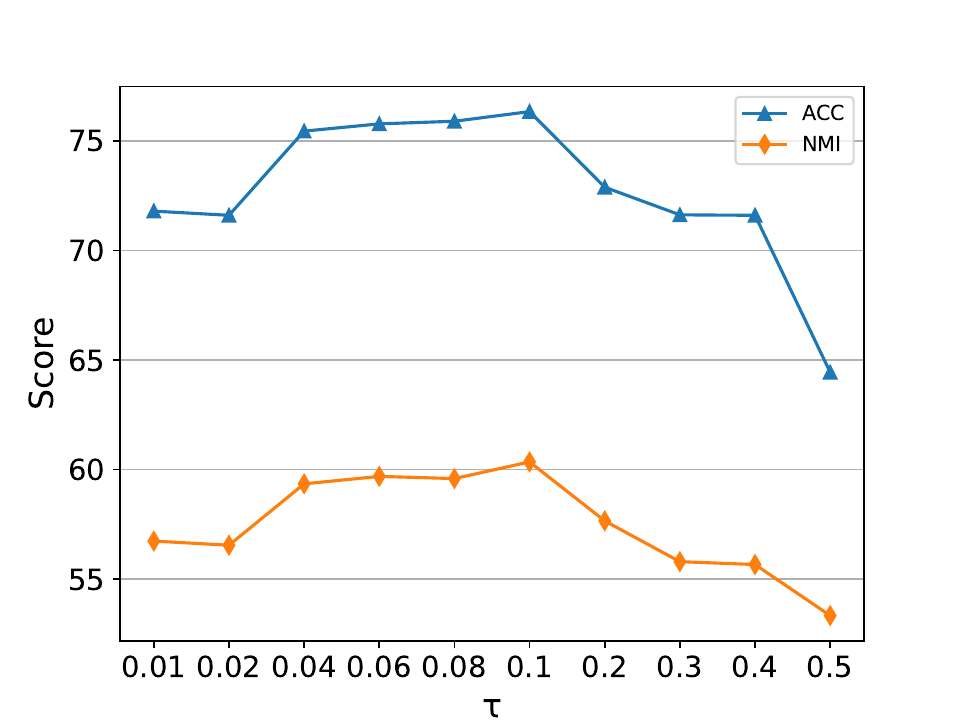}}  
    \subfigure[WikiCS]{\includegraphics[width=0.49\linewidth]{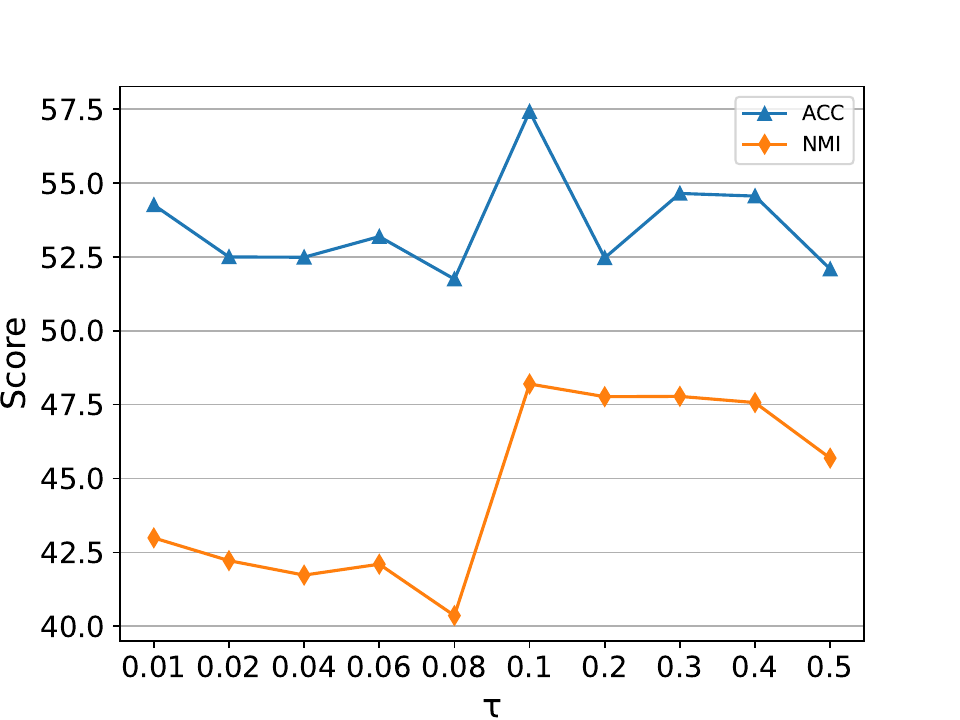}}
    \subfigure[Photo]{\includegraphics[width=0.49\linewidth]{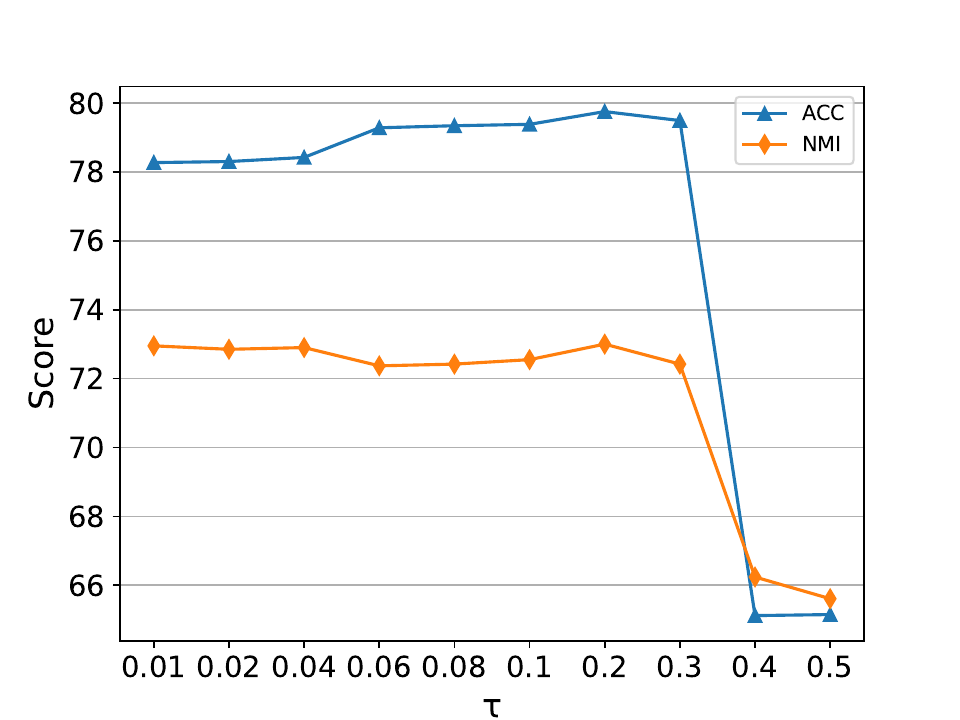}}
    \subfigure[CoauthorCS]{\includegraphics[width=0.49\linewidth]{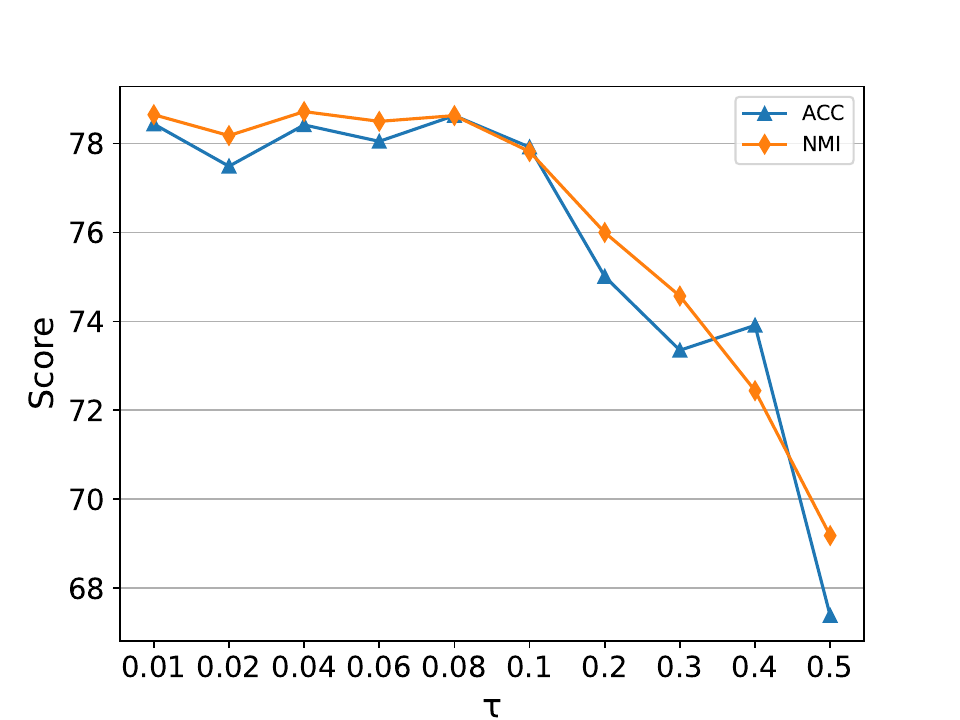}}
    \caption{Impact of the temperature $\tau$ on Cora, WikiCS, Photo and CoauthorCS.}
    \label{Fig: param_tau}
\end{figure}

\textbf{Impact of the temperature $\tau$.} We further explore the impact of the temperature $\tau$ on NS4GC. Figure \ref{Fig: param_tau} illustrates the ACC and NMI metrics for four datasets across different values of $\tau = \{0.01, 0.02, 0.04, 0.06, 0.08, 0.1, 0.2, 0.3, 0.4, 0.5 \}$. Notably,a value that is too small (e.g., 0.01) leads to vanishing gradients in the sigmoid activation in the Eq. \ref{Eq: Sparsity Penalty}, while a too large $\tau$ (e.g., 0.5) fails to enforce $\hat{w}_{ij}$ to be binary. And it is observed that setting $\tau$ to 0.1 is sufficient for all datasets.

\begin{figure}[!ht]
    \centering
    \subfigure[Cora]{\includegraphics[width=0.49\linewidth]{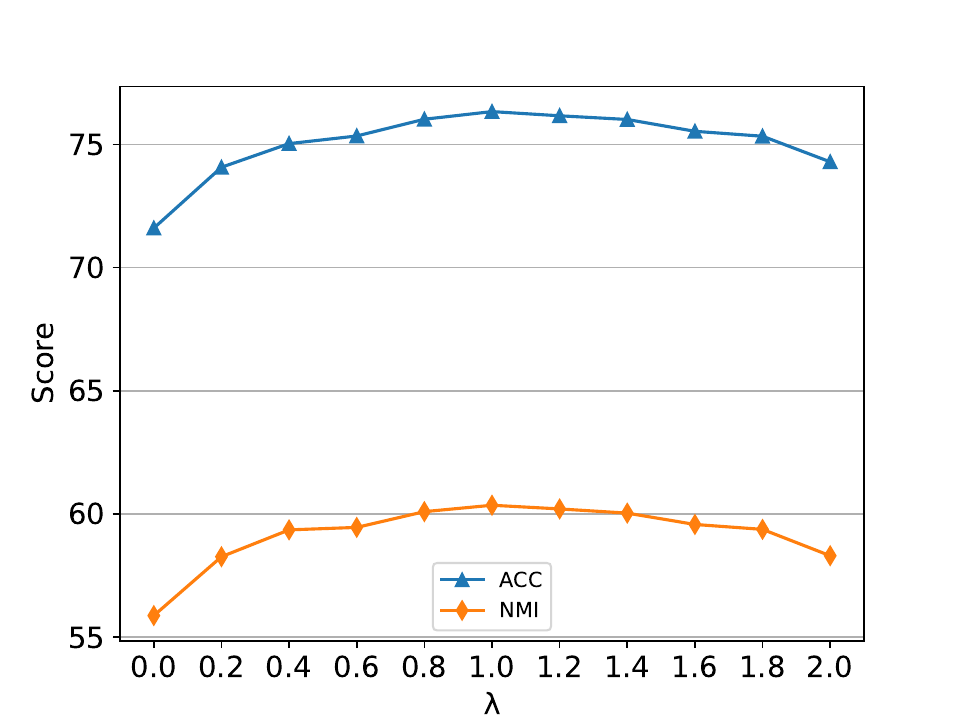}}  
    \subfigure[WikiCS]{\includegraphics[width=0.49\linewidth]{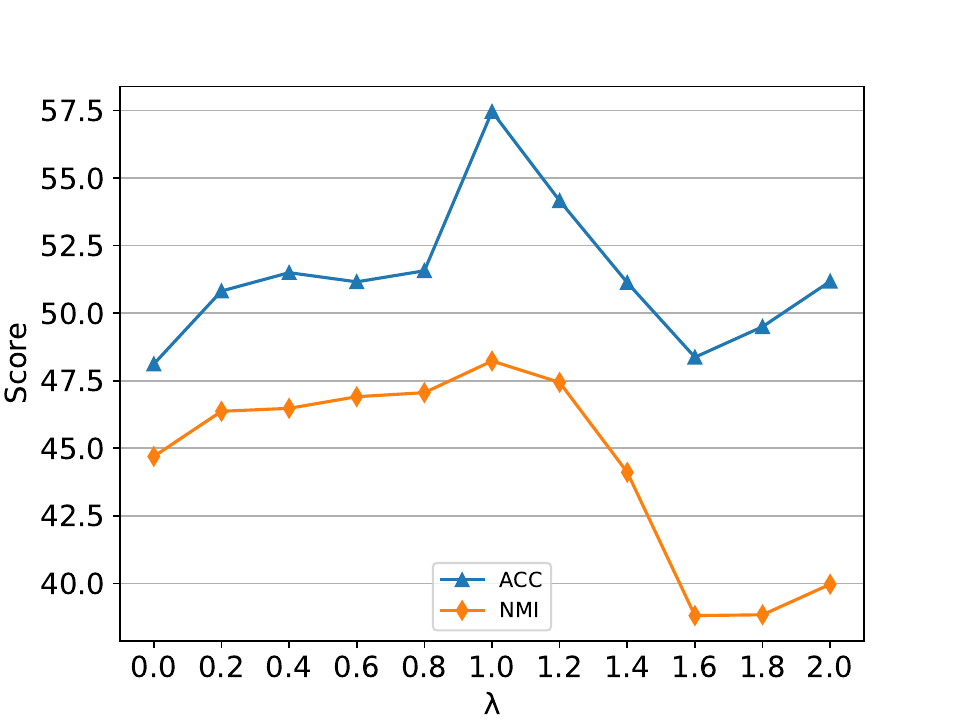}}
    \subfigure[Photo]{\includegraphics[width=0.49\linewidth]{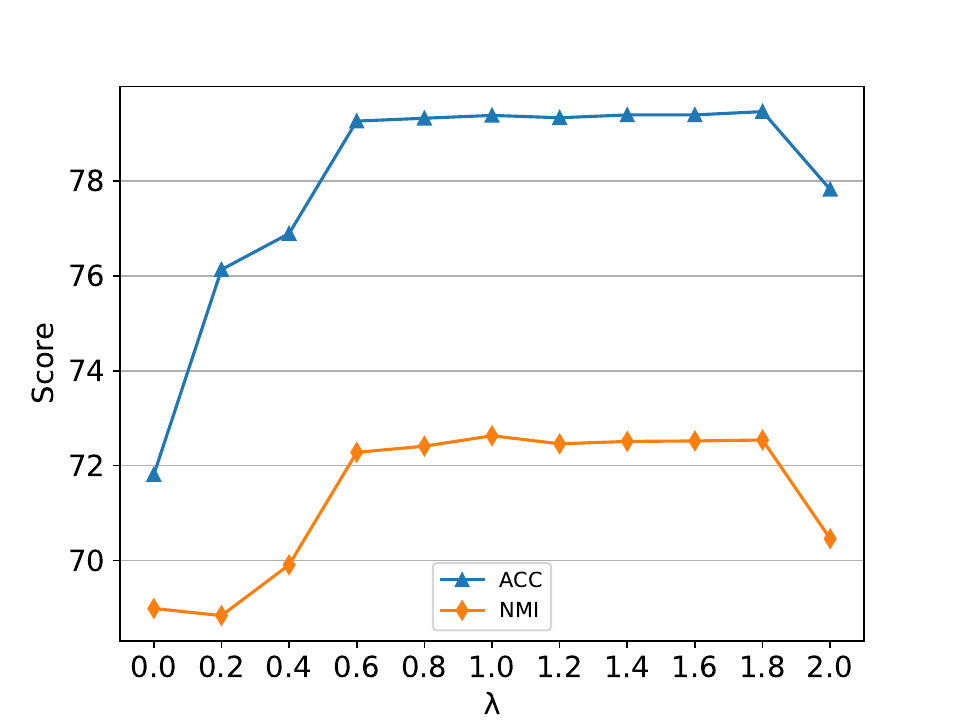}}
    \subfigure[CoauthorCS]{\includegraphics[width=0.49\linewidth]{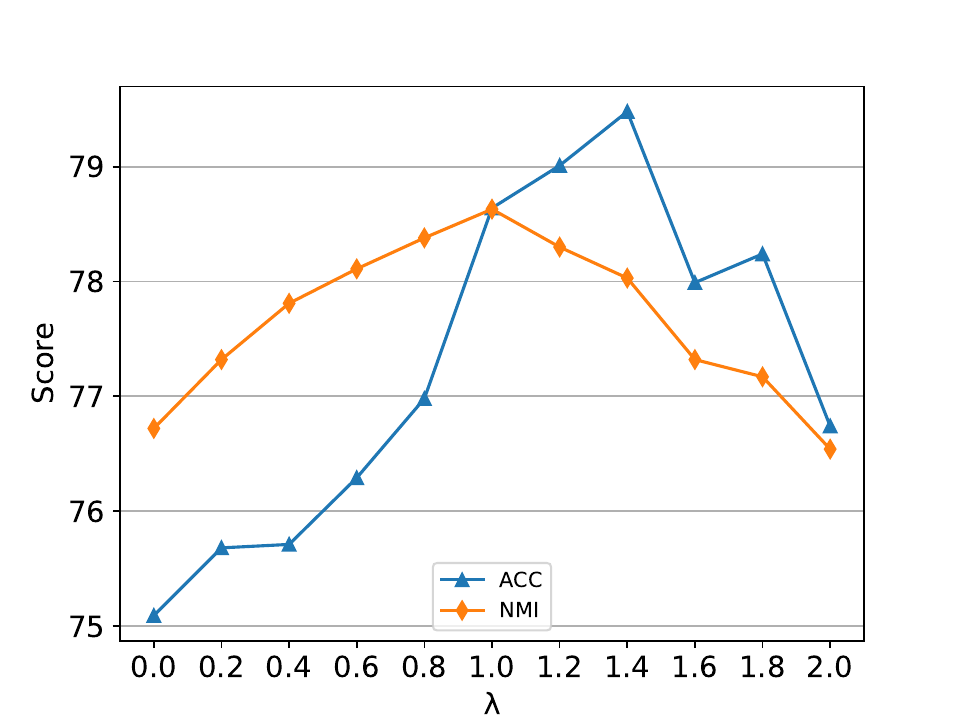}}
    \caption{Impact of the trade-off $\lambda$ on Cora, WikiCS, Photo and CoauthorCS.}
    \label{Fig: param_lam}
\end{figure}

\textbf{Impact of the trade-off $\lambda$.} We additionally investigate the impact of the intensity of the node-neighbor alignment term on performance by varying the trade-off hyperparameter $\lambda$ from 0.0 to 2.0 in increments of 0.2. Figure \ref{Fig: param_lam} visually presents the ACC and NMI metrics on Cora, WikiCS, Photo, and CoauthorCS. It is observed that, initially, increasing the intensity of the node-neighbor alignment term improves performance, but excessive intensity results in a performance decline. In general, NS4GC exhibits robustness to variations in $\lambda$. And we recommend selecting a $\lambda$ value of $1$ for all datasets.

\begin{figure}[!ht]
    \centering
    \subfigure[Cora]{\includegraphics[width=0.49\linewidth]{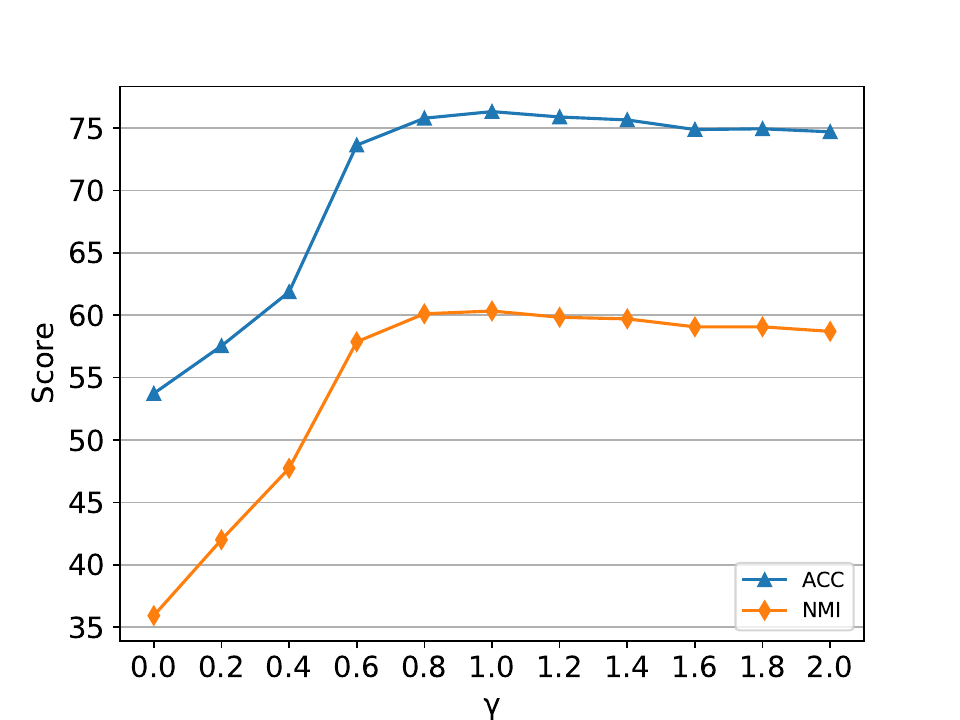}}  
    \subfigure[WikiCS]{\includegraphics[width=0.49\linewidth]{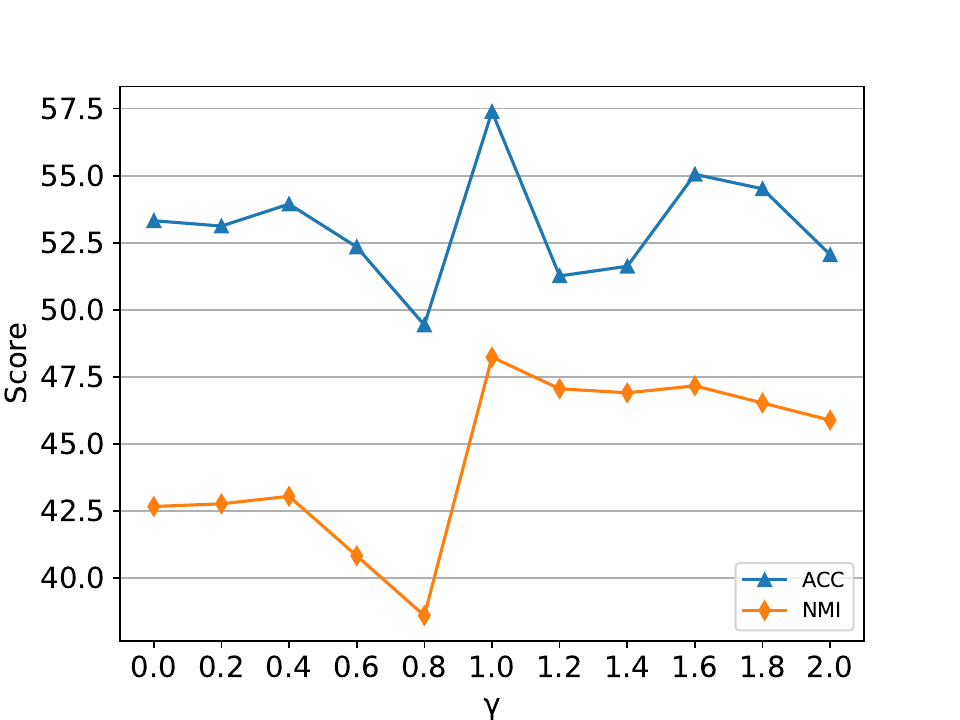}}
    \subfigure[Photo]{\includegraphics[width=0.49\linewidth]{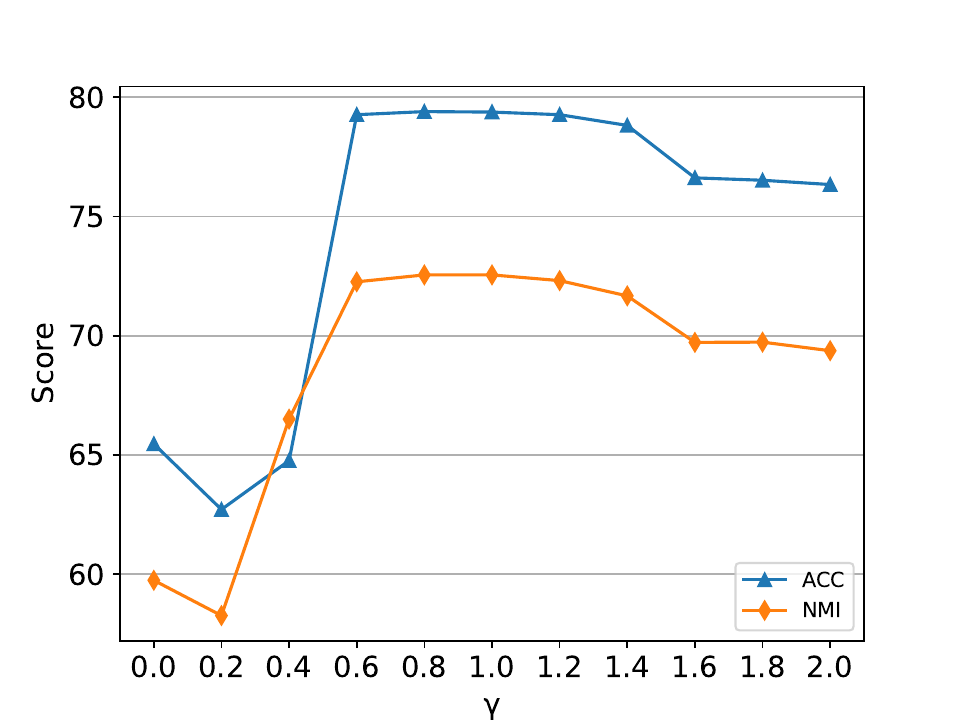}}
    \subfigure[CoauthorCS]{\includegraphics[width=0.49\linewidth]{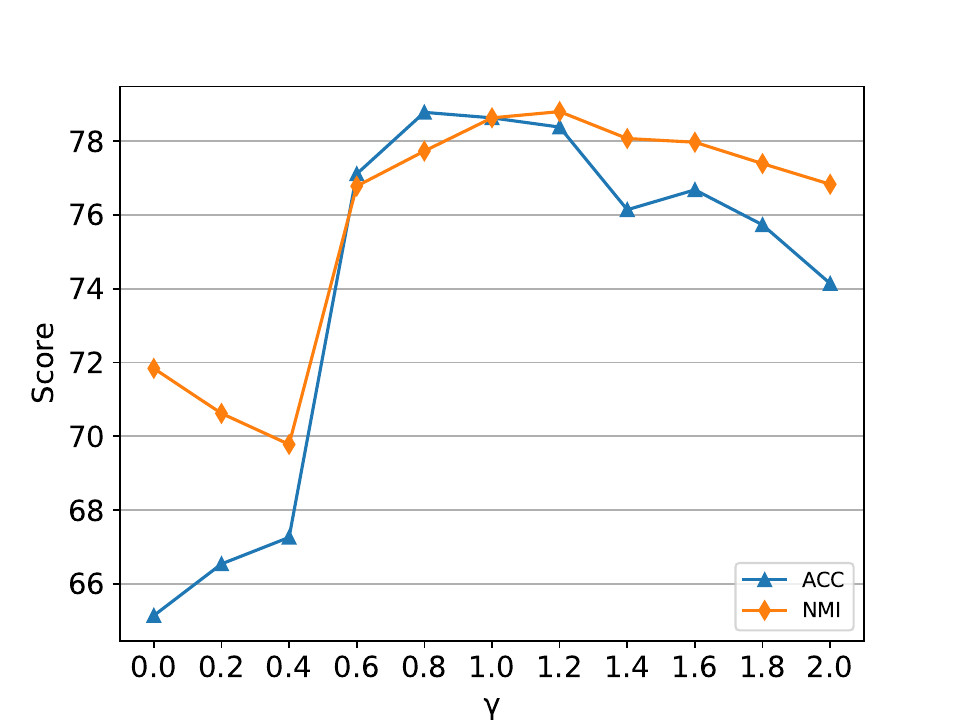}}
    \caption{Impact of the trade-off $\gamma$ on Cora, WikiCS, Photo and CoauthorCS.}
    \label{Fig: param_gam}
\end{figure}

\textbf{Impact of the trade-off $\gamma$.} We also investigate the impact of the intensity of the sparsity penalty term on performance by varying the trade-off hyperparameter $\gamma$ from 0.0 to 2.0 in increments of 0.2. Figure \ref{Fig: param_gam} visually presents the ACC and NMI metrics for four datasets. It is evident that neither extremely small nor excessively large values of $\gamma$ yield optimal performance. This observation supports the advantage of our proposed NSGC over GRACE and CCASSG, as these methods can be seen as extreme examples that employ the sparsity penalty on all non-diagonal elements with either a overly large $\gamma$ or a very small $\gamma$, resulting in under-sparse or over-sparse node similarity matrices. In contrast, by carefully selecting an appropriate value of $\gamma$, such as $1$, our proposed NS4GC can learn an approximately ideal node similarity matrix within the representation space, thereby achieving more clustering-friendly representations. 

\begin{figure}[!ht]
    \centering
    \subfigure[Cora: ACC]{\includegraphics[width=0.49\linewidth]{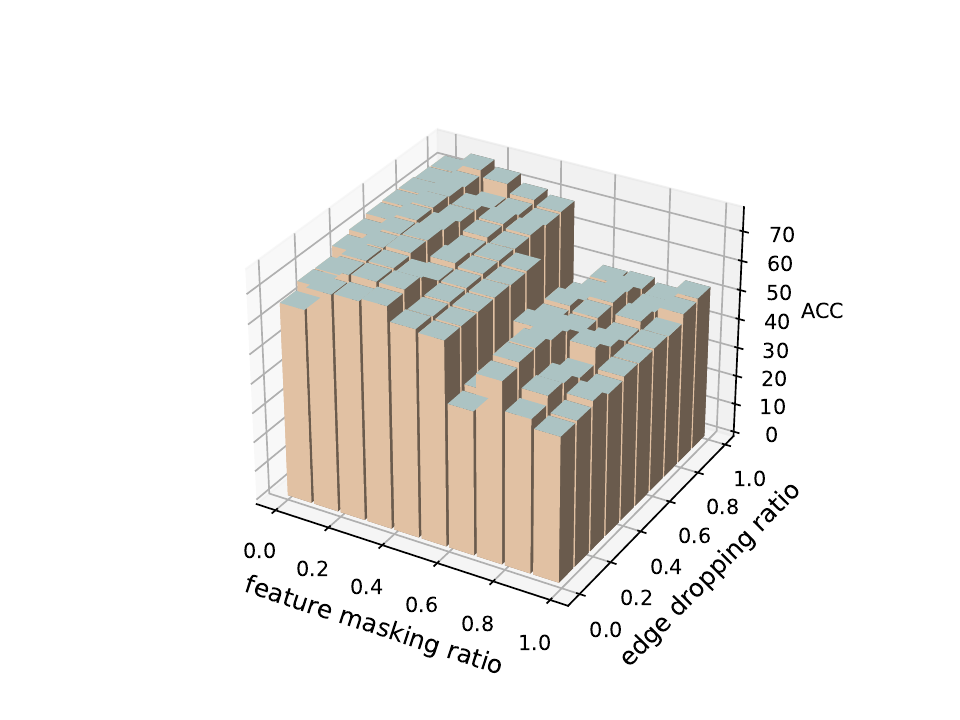}} 
    \subfigure[Cora: NMI]{\includegraphics[width=0.49\linewidth]{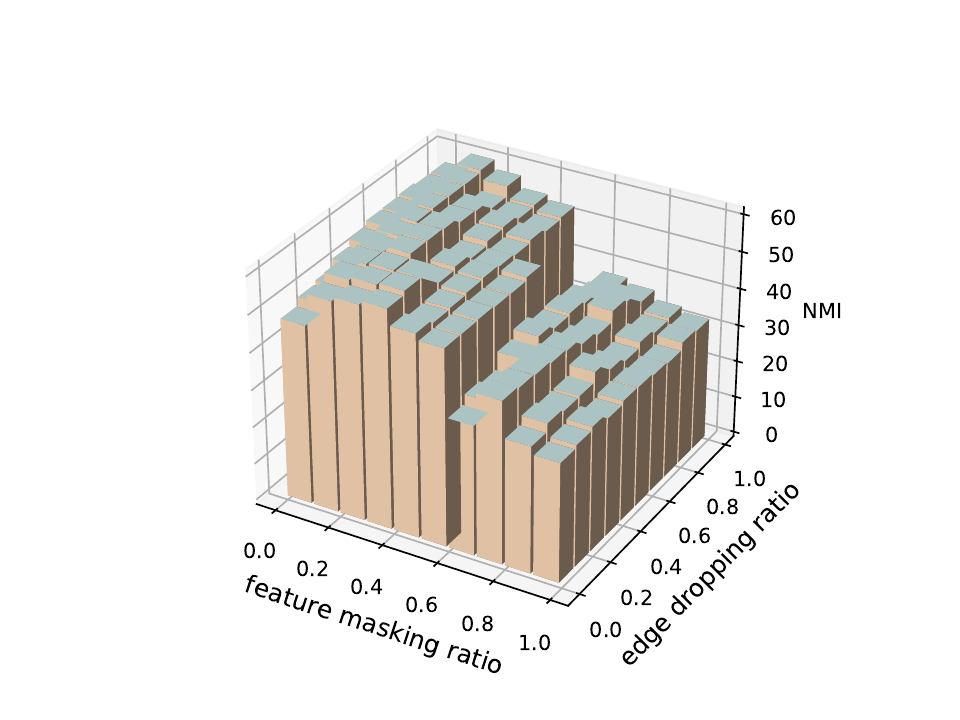}}
    
    \subfigure[Citeseer: ACC]{\includegraphics[width=0.49\linewidth]{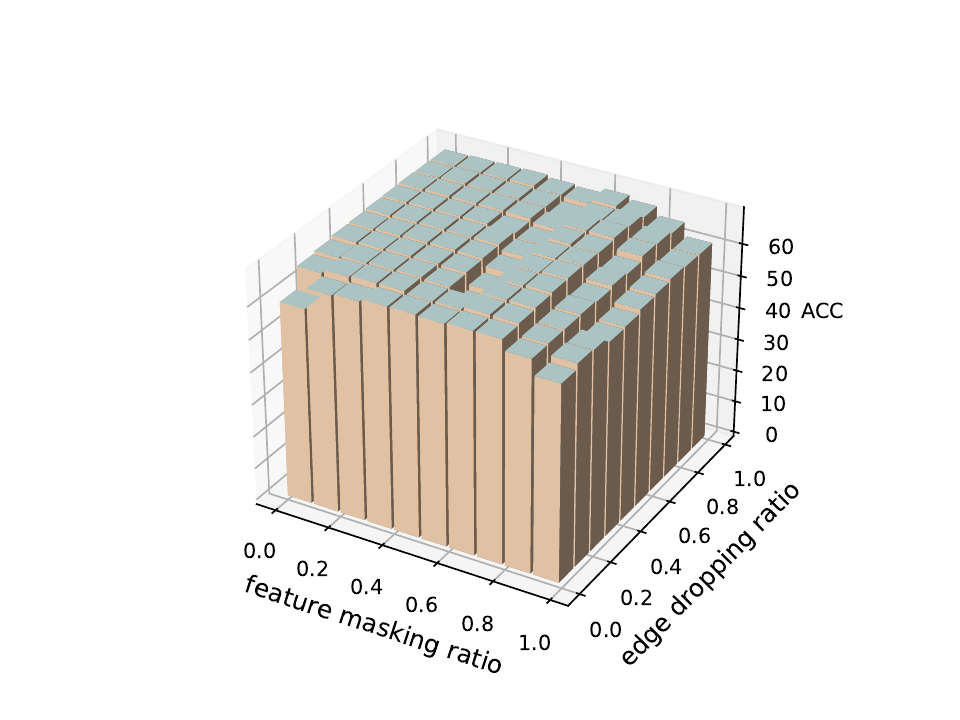}}
    \subfigure[Citeseer: NMI]{\includegraphics[width=0.49\linewidth]{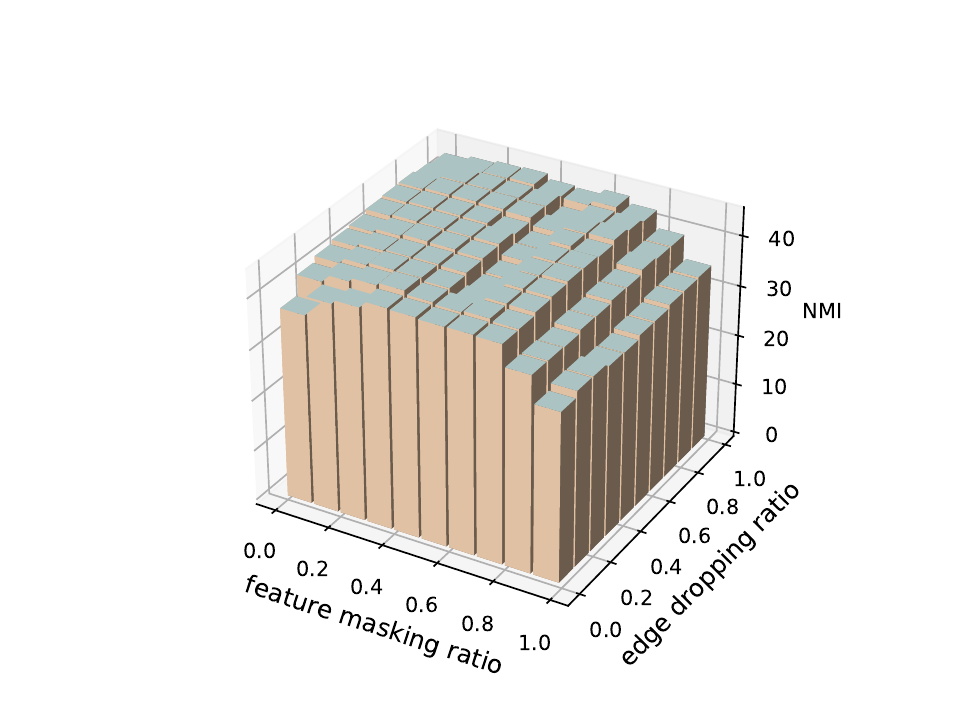}}
    
    \subfigure[Photo: ACC]{\includegraphics[width=0.49\linewidth]{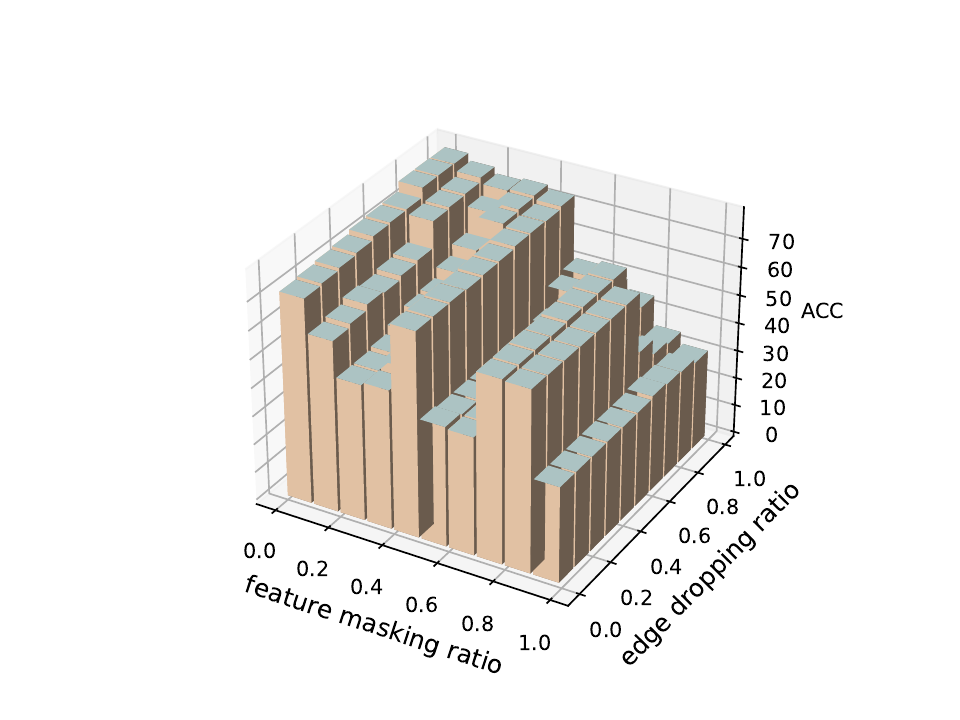}}    
    \subfigure[Photo: NMI]{\includegraphics[width=0.49\linewidth]{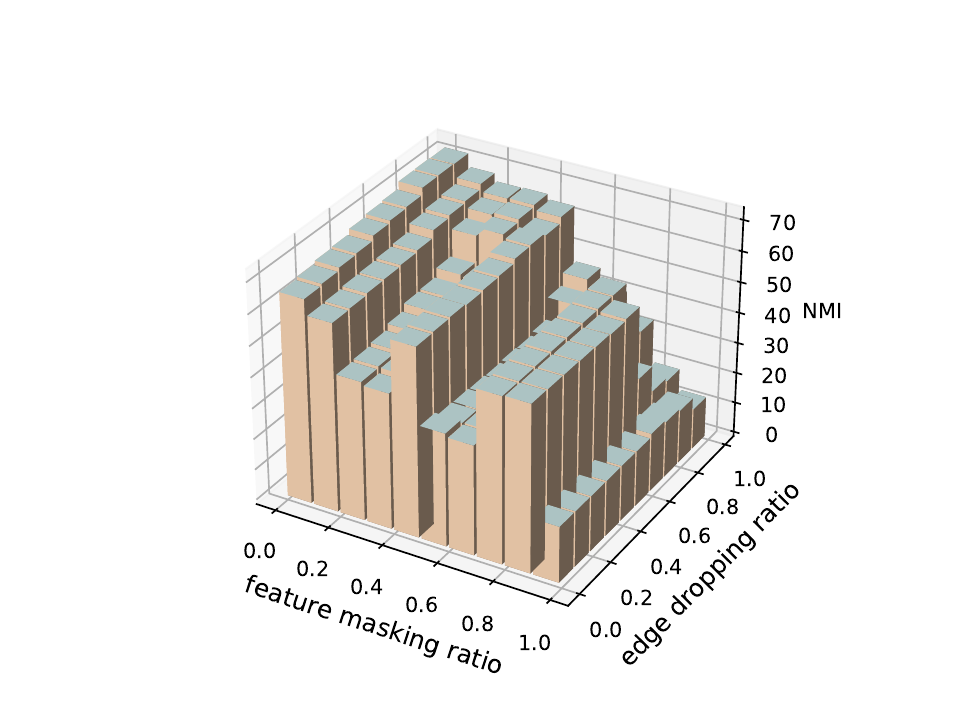}}
    \caption{Impact of different augmentation intensity on Cora, Citeseer and Photo.}
    \label{Fig: param_pdpm}
\end{figure}

\textbf{Impact of the augmentation intensity $p_d$ and $p_m$.} We additionally investigate the impact of augmentation intensity, specifically $p_{d1}, p_{d2}, p_{m1}, p_{m2}$, on node clustering. Our experiments encompass Cora, Citeseer, and Photo datasets, with variations of these parameters from 0.0 to 0.9. To simplify visualization, we set $p_{d1} = p_{d2}$ and $p_{m1} = p_{m2}$. All other hyperparameters remain consistent with the previously described settings. The results, depicted in Figure \ref{Fig: param_pdpm}, indicate that our method is more sensitive to augmentation in features compared to that in graph structure. Overall, our method exhibits robustness to augmentation intensity: by maintaining the feature masking ratio and the edge dropping ratio within the appropriate range, our method achieves impressive and competitive performance.  However, it is still very important to select a proper augmentation intensity as well as augmentation method to learn more clustering-friendly representations.

\section{Conclusion}
In this work, we explore contrastive graph clustering from the perspective of the node similarity matrix, recognizing the fundamental role of similarity measures in clustering. Our analysis reveals that current contrastive graph clustering methods inadequately explore node-wise similarity. They either assume the node similarity matrix within the representation space to be an identity matrix or lack explicit constraints on off-diagonal entries, potentially leading to collapsed representations. To address these shortcomings, we propose NS4GC, a novel framework that employs a node-neighbor alignment and a semantic-aware sparsification to construct an approximately ideal node similarity matrix. This matrix ensures representations of semantically similar nodes are positioned closely within the representation space, resulting in clustering-friendly representations. Extensive experiments conducted on eight benchmark datasets demonstrate the effectiveness of our proposed method.

\noindent \textbf{Limitations.} We note that the proposed method NS4GC has a limitation during deployment, which relies on the assumption about the underlying graph structure and cluster distribution. Namely, homophilous graphs (i.e., the connected nodes are more likely to belong to the same cluster) are more suited to NS4GC. However, if the graph is non-homophily, using the node-neighbor alignment to encourage two connected nodes similar may not help node clustering.

\section*{Acknowledgment}
This work is partially supported by the National Key Research and Development Program of China (2021YFB1715600), the National Natural Science Foundation of China (62306137), the Australian Research Council under the streams of Future Fellowship (FT210100624), Linkage Project (LP230200892), and Discovery Project (DP240101108).

\bibliographystyle{IEEEtran}
\bibliography{references}

\vspace{-13mm}
\begin{IEEEbiography}[{\includegraphics[width=1in,height=1.25in,clip,keepaspectratio]{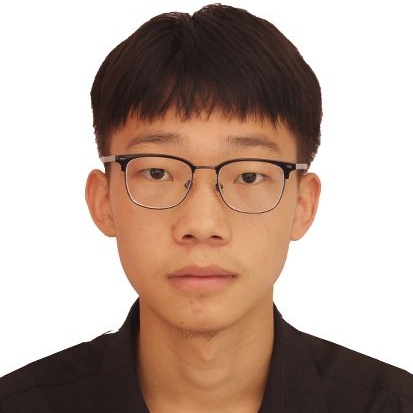}}]{Yunhui~Liu} is working toward the Ph.D. degree with the Software Institute, Nanjing University, China. His research interests include graph machine learning and self-supervised learning.
\end{IEEEbiography}

\vspace{-16mm}
\begin{IEEEbiography}[{\includegraphics[width=1in,height=1.25in,clip,keepaspectratio]{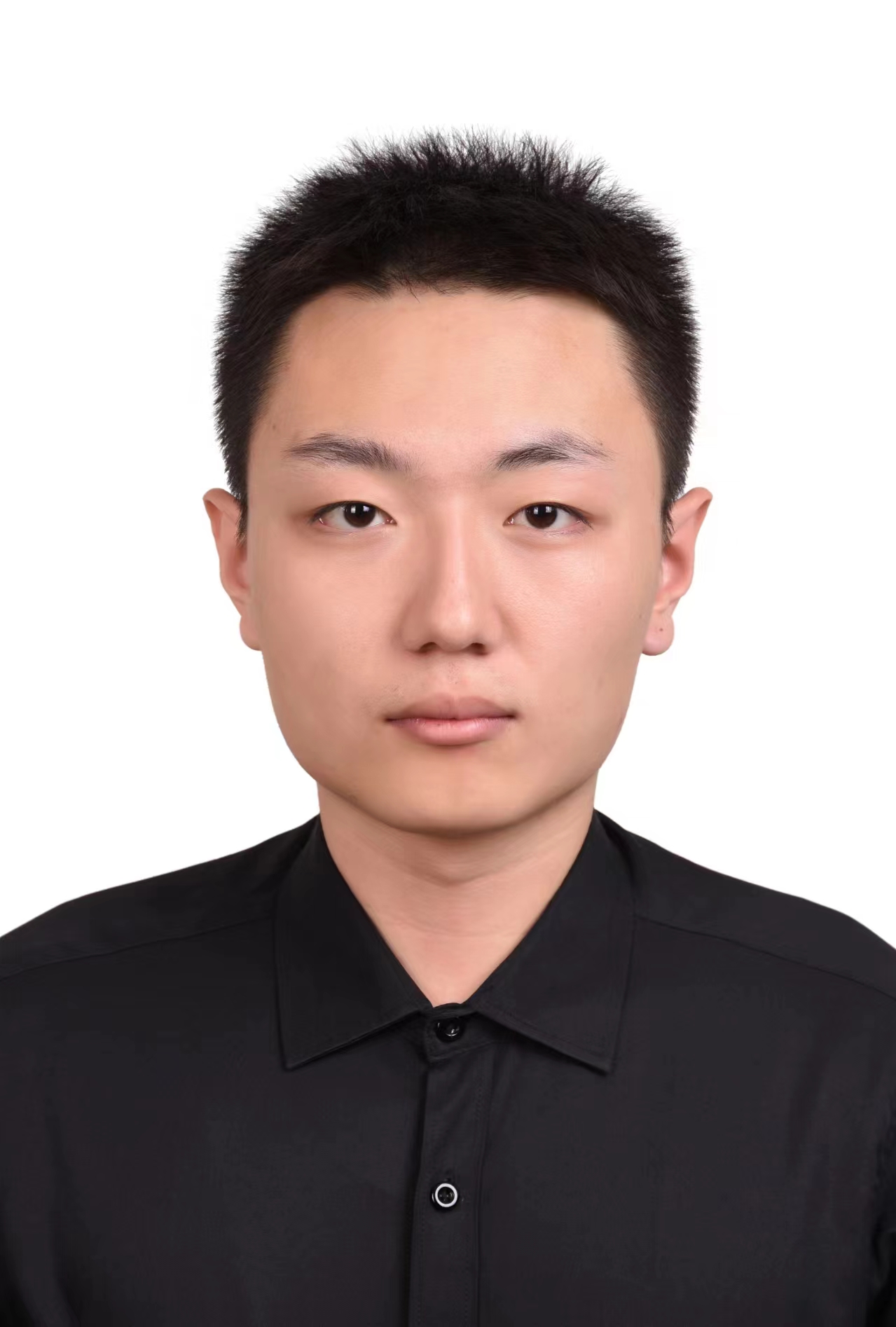}}]{Xinyi~Gao} received his B.S. and M.S. degrees in Information and Communications Engineering from Xi'an Jiaotong University, China. Currently, he is a Ph.D. candidate at the School of Electrical Engineering and Computer Science, the University of Queensland. His research interests include data mining, graph representation learning, and multimedia signal processing.
\end{IEEEbiography}

\vspace{-15mm}
\begin{IEEEbiography}[{\includegraphics[width=1in,height=1.25in,clip,keepaspectratio]{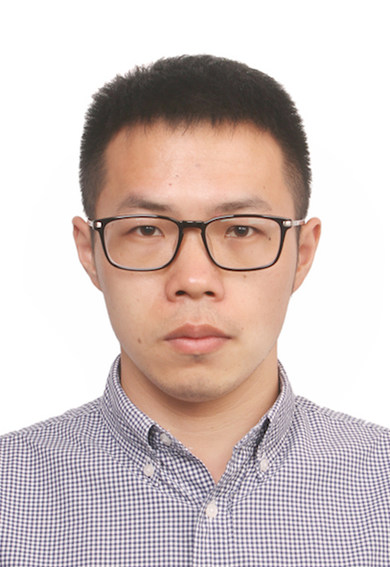}}]{Tieke~He} received the B.E. and Ph.D. degrees in software engineering from the Software Institute, Nanjing University, Jiangsu, China. He is currently an Associate Professor with the Software Institute, Nanjing University. His research interests lie in intelligent software engineering, knowledge graph, and question answering.
\end{IEEEbiography}

\vspace{-15mm}
\begin{IEEEbiography}[{\includegraphics[width=1in,height=1.25in,clip,keepaspectratio]{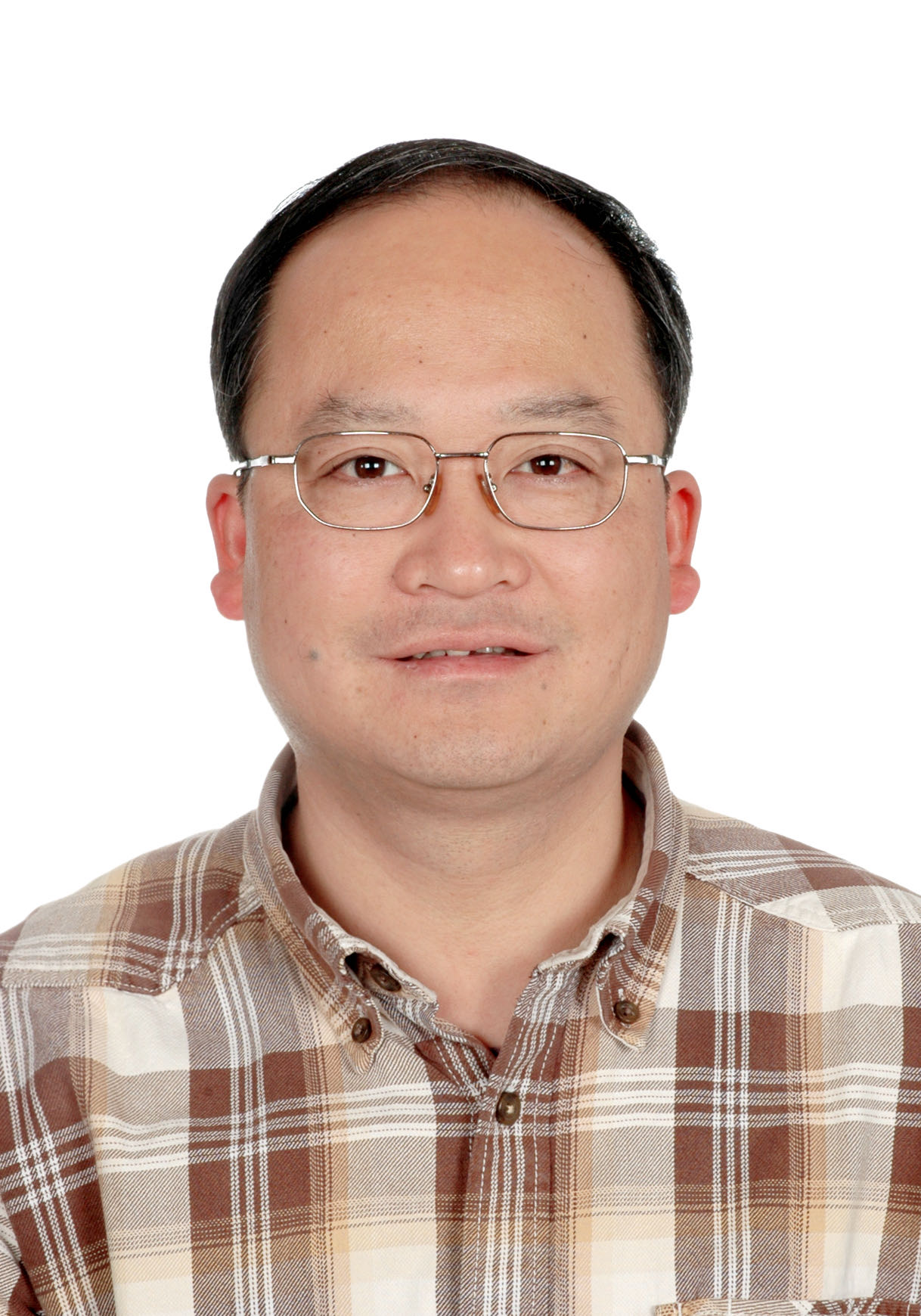}}]{Tao~Zheng} is a full professor with the Software Institute, Nanjing University. His research interests include formal methods, programming language, natural language processing, and knowledge graph.
\end{IEEEbiography}

\vspace{-13mm}
\begin{IEEEbiography}[{\includegraphics[width=1in,height=1.25in,clip,keepaspectratio]{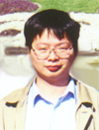}}]{Jianhua~Zhao} received the B.E., M.E., and Ph.D. degrees in computer science from Nanjing University, Nanjing, China, in 1993, 1996, and 1999, respectively. He is a professor with Nanjing University. His research interests include formal support for design and analysis of systems, software verification.
\end{IEEEbiography}

\vspace{-15mm}
\begin{IEEEbiography}[{\includegraphics[width=1in,height=1.25in,clip,keepaspectratio]{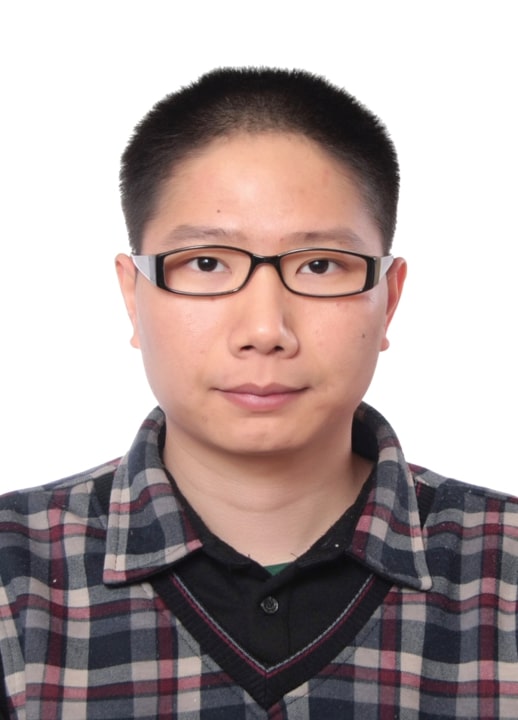}}]{Hongzhi~Yin} received a Ph.D. degree in computer science from Peking University, in 2014. He works as an ARC Future Fellow, Full Professor, and Director of the Responsible Big Data Intelligence Lab (RBDI) at The University of Queensland, Australia. He has made notable contributions to predictive analytics, recommendation systems, graph learning, and decentralized and edge intelligence. He has published over 300 papers with an H-index of 73 and received 8 best paper awards/runner-ups at the top conferences.
\end{IEEEbiography}

\end{document}